\documentclass[10pt,twocolumn,letterpaper]{article}

\usepackage{cvpr}              % To produce the CAMERA-READY version
\usepackage{graphicx}
\usepackage{amsmath}
\usepackage{amssymb}
\usepackage{booktabs}

\usepackage[pagebackref,breaklinks,colorlinks]{hyperref}

\usepackage[capitalize]{cleveref}
\crefname{section}{Sec.}{Secs.}
\Crefname{section}{Section}{Sections}
\Crefname{table}{Table}{Tables}
\crefname{table}{Tab.}{Tabs.}

\usepackage{times}  % DO NOT CHANGE THIS
\usepackage{helvet}  % DO NOT CHANGE THIS
\usepackage{courier}  % DO NOT CHANGE THIS
\usepackage[hyphens]{url}  % DO NOT CHANGE THIS
\usepackage{graphicx} % DO NOT CHANGE THIS
\usepackage{caption} % DO NOT CHANGE THIS AND DO NOT ADD ANY OPTIONS TO IT
\DeclareCaptionStyle{ruled}{labelfont=normalfont,labelsep=colon,strut=off} % DO NOT CHANGE THIS
\usepackage{algorithm}
\usepackage{algorithmic}

\usepackage{times}
\usepackage{epsfig}
\usepackage{amsmath}
\usepackage{amssymb}

\usepackage{enumitem}
\usepackage{multicol}
\usepackage{lineno}
\usepackage{bm}
\usepackage{ragged2e}
\usepackage{caption}
\usepackage[para,online,flushleft]{threeparttable}
\usepackage{booktabs}
\usepackage{multicol}
\usepackage{multirow}
\usepackage{xpatch}
\usepackage{nicefrac}  % compact symbols for 1/2, etc.
\usepackage{tabularx}
\usepackage{colortbl}
\usepackage{newfloat}
\usepackage{listings}
\floatstyle{ruled}
\newfloat{listing}{tb}{lst}{}
\floatname{listing}{Listing}

\usepackage{array, multirow}
\newcolumntype{C}[1]{>{\centering\let\newline\\\arraybackslash\hspace{0pt}}m{#1}}

\newcommand{\squeeze}{\textsc{Squeeze}}
\newcommand{\osplit}{\textsc{Split}}
\newcommand{\neuralNet}{\textsc{NN}}

\newcommand{\sflow}{\textsc{StepOfFlow}}

\newcommand{\parsection}[1]{\vspace{0.5mm}\noindent\textbf{#1.}~}

\pdfoutput=1

\def\cvprPaperID{6058} % *** Enter the CVPR Paper ID here
\def\confName{CVPR}
\def\confYear{2022}

\begin{document}

\title{Generative Flows with Invertible Attentions}

\author{Rhea Sanjay Sukthanker\textsuperscript{\rm 1}, Zhiwu Huang\textsuperscript{\rm 1,2}, Suryansh Kumar\textsuperscript{\rm 1}, Radu Timofte\textsuperscript{\rm 1}, Luc Van Gool\textsuperscript{\rm 1,3}\\
	\textsuperscript{\rm 1}CVL, ETH Z\"urich, Switzerland \quad
	\textsuperscript{\rm 2}SAVG, SMU, Singapore \quad
	\textsuperscript{\rm 3}PSI, KU Leuven, Belgium\\
	{\tt\small srhea@alumni.ethz.ch \quad \{zhiwu.huang,sukumar,radu.timofte,vangool\}@vision.ee.ethz.ch}
	% \and
}

\maketitle

%%%%%%%%% ABSTRACT
\begin{abstract}
	
Flow-based generative models have shown an excellent ability to explicitly learn the probability density function of data via a sequence of invertible transformations. Yet, learning attentions in generative flows remains understudied, while it has made breakthroughs in other domains. To fill the gap, this paper introduces two types of invertible attention mechanisms, \ie, map-based and transformer-based attentions, for both unconditional and conditional generative flows.
The key idea is to exploit a masked scheme of these two attentions to learn long-range data dependencies in the context of generative flows.
The masked scheme allows for invertible attention modules with tractable Jacobian determinants, enabling its seamless integration at any positions of the flow-based models. The proposed attention mechanisms lead to more efficient generative flows, due to their capability of modeling the long-term data dependencies. Evaluation on multiple image synthesis tasks shows that the proposed attention flows result in efficient models and compare favorably against the state-of-the-art unconditional and conditional generative flows.

\end{abstract}
%%%%%%%%% BODY TEXT
\section{Introduction}
Deep generative models have shown their capability to model complex real-world datasets for various applications, such as image synthesis \cite{goodfellow2014generative,kingma2013auto,dinh2015nice,tian2020off,shahbazi2021efficient}, image super-resolution \cite{ledig2017photo,wang2018esrgan}, facial manipulation \cite{huang2017beyond,choi2018stargan,pumarola2018ganimation,d2021ganmut}, autonomous driving \cite{wang2020improving,zhang2020weakly}, and others.
The widely studied modern generative models include generative adversarial nets (GANs) \cite{goodfellow2014generative,brock2018large,wu2018wasserstein,karras2019style}, variational autoencoders (VAEs) \cite{kingma2013auto,makhzani2015adversarial,tolstikhin2018wasserstein,wu2019sliced}, 
autoregressive models \cite{van2016conditional, van2016pixel} and flow-based models \cite{dinh2017density,dinh2015nice,kingma2018glow}.
The GAN models implicitly learn the data distribution to produce samples by transforming a noise distribution into the desired space, where the generated data can approximate the real data distribution. On the other hand, VAEs optimize a lower bound on the data's log-likelihood, leading to a suitable approximation of the actual data distribution. Although these two models have achieved great success, neither provides exact data likelihood.

\begin{figure}[t]
\begin{center}
\includegraphics[width=1.0\linewidth]{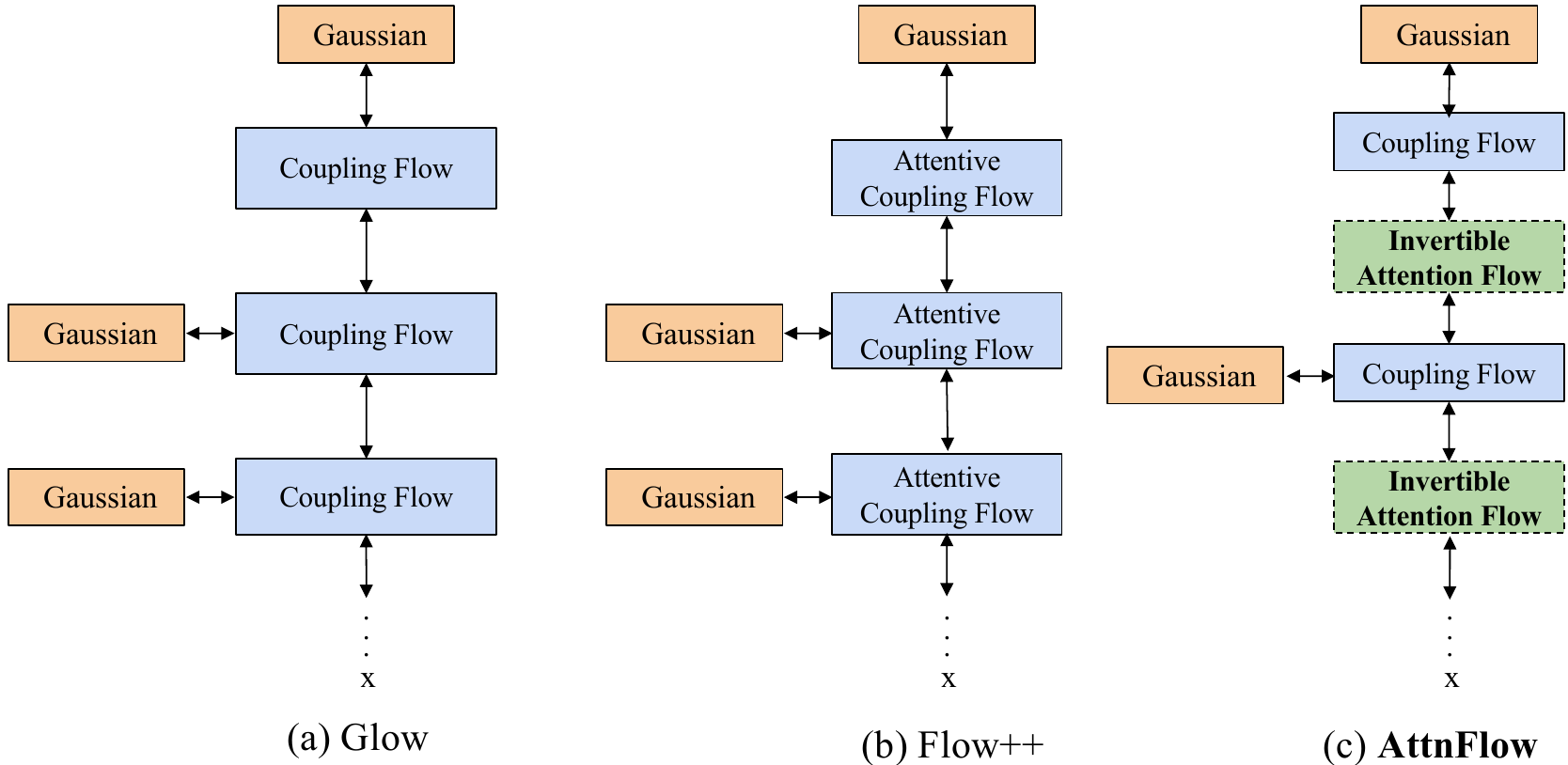}
\end{center}
\vspace{-0.2cm}
  \caption{{\small Conceptual comparison of the proposed AttnFlow against two representative generative flows, \ie, (a) Glow \cite{kingma2018glow} and (b) Flow++ \cite{ho2019flow++}. Based on \cite{kingma2018glow}, Flow++ introduces the conventional attention mechanism to model short-term dependencies within one split of each feature map in the context of coupling layers. In contrast, the proposed AttnFlow (shown in (c)) further introduces invertible attention mechanisms that can be introduced at
  any flow positions to learn long-term correlations.}}
   \vspace{-0.2cm}
\label{fig:concept}
\end{figure}

Autoregressive models \cite{domke2008killed,van2016conditional,van2016pixel} and flow-based generative models \cite{dinh2015nice,dinh2017density,kingma2018glow} optimize the exact log-likelihood of real data. Despite autoregressive models' better performance on density estimation benchmarks, its sequential property results in non-trivial parallelization. In contrast, the flow-based generative models are conceptually attractive due to tractable log-likelihood, exact latent-variable inference, and parallelizability of both training and synthesis. Notably, they allow exact inference of the actual data log-likelihood via normalizing flow. As shown in Fig.\ref{fig:concept}(a), the normalizing flow model transforms a simple distribution into a complex one by applying a sequence of invertible transformation functions, which leads to an excellent mechanism of simultaneous exact log-likelihood optimization and latent-variable inference. However, due to efficiency constraints in their network designs, most models require several flow layers to approximate non-linear long-range data dependencies to get globally coherent samples. To overcome this drawback \ie, modeling dependencies efficiently over normalizing flows is the key, and presently one of the most sought-after problems \cite{ho2019flow++, mahajan2020normalizing}.

To efficiently model data dependencies in the flow-based generative models, one may opt to combine multi-scale autoregressive priors \cite{mahajan2020normalizing}. By comparison, exploiting attention mechanisms has emerged as a remarkable way to model such dependencies in deep neural networks. It imitates human brain actions of selectively concentrating on a few relevant information while ignoring less correlated ones. Traditional self-attention mechanisms like \cite{vaswani2017attention,zhang2019self,wang2020attentionnas} exhibit a good balance between the ability to model range dependencies and the computational and statistical efficiency. In general, the self-attention modules measure the response at a point as a weighted sum of the features at all points, where the attention weights are computed at a small computational cost. Although \cite{ho2019flow++} recently applied the conventional attention directly as a dependent component in the coupling layer (Fig.\ref{fig:concept}(b)), it models dependency within a short-range (\ie, one split of each flow feature map) of the coupling layer. To our knowledge, efficient modeling of data dependencies over normalizing flows is understudied. A natural solution is to exploit new attention mechanisms to learn correlations of the feature maps at any positions of the flow-based models. However, it is generally non-trivial to achieve that goal of exploiting new attention modules as independent flow layers. Concretely, such attentions should maintain the invertibility with tractable Jacobian determinants in the flows. %flow-based generative models.

In this paper, we propose invertible attentions for flow (AttnFlow) models to reliably and efficiently model network data dependencies that can be introduced at any positions of the flow-based models (along the entire flow feature maps, see Fig.\ref{fig:concept}(c)). The key idea is to exploit a masked attention learning scheme to allow for intertible attention learning for normalizing flow based generative models. In addition, the proposed masked attention scheme facilitates tractable Jacobian determinants and hence can be integrated seamlessly into any generative flow models. Particularly, we exploit two different invertible attention mechanisms to encode the various types of correlations respectively on the flow feature maps. The two proposed attention mechanisms are \textit{\textbf{(i)} invertible map-based (iMap) attention} that directly models the importance of each position in the attention dimension of the flow feature maps, 
\textit{\textbf{(ii)} invertible transformer-based (iTrans) attention} 
that explicitly models the second-order interactions among distant positions in the attention dimension. Since the proposed two invertible attention modules explicitly model the dependencies of flow feature maps, it further enhances a flow-based model's efficiency to represent the deep network dependencies. To show the superiority of our approach, we evaluated the introduced attention models in the context of both unconditional and conditional normalizing flow-based generative models for multiple image synthesis tasks.
\section{Related Work}
\label{sec2}
\parsection{Generative Flows}
Early flow-based generative models like \cite{dinh2015nice,kingma2016improved,dinh2017density} are introduced for exact inference of real data log-likelihood. They are generally constructed by a sequence of invertible transformations to map a base distribution to a complex one.

%\begin{table}[htbp]
%	\centering
%	\begin{tabular}{|c|c|c|c|p{1cm}p{1cm}p{1cm}p{1cm}p{1cm}p{1cm}p{1cm}|}
%		\hline
%		A & B & C & D & \multicolumn{7}{|c|}{F}  \\ \hline
%		\multirow{ 2}{*}{1} & 0 & 6 & 230 & 35 & 40 & 55 & 25 & 40 & 35 & \\
%		& 1 & 5 & 195 & 25 & 50 & 35 & 40 & 45 &  &  \\ \hline
%	\end{tabular}
%	\caption{A test caption}
%	\label{table2}
%\end{table}

\begin{table*}[t]
    \centering
	\footnotesize
    \begin{tabular}{|m{0.12\textwidth}|p{0.32\textwidth}||m{0.12\textwidth}|p{0.32\textwidth}|}
    \hline
    \rowcolor[gray]{0.85}
\textbf{Layer}  & \textbf{Function} & \textbf{Layer}   & \textbf{Function} \\ \hline
Actnorm & $\forall i,j: \mathbf{y}_{i,j} = \mathbf{s} \odot \mathbf{x}_{i,j} + \mathbf{b}$ & Invertible $1 \times 1$ Convolution  & $\forall i,j: \mathbf{y}_{i,j} =\mathbf{W} \mathbf{x}_{i,j}$\\
\hline
 \multirow{4}{*} {Affine Coupling} 
 & $\mathbf{x}_a, \mathbf{x}_b = \osplit(\mathbf{x})$  & \multirow{4}{*} {\shortstack[l]{Mixture Affine \\ Coupling}}  &  $\mathbf{x}_a, \mathbf{x}_b = \osplit(\mathbf{x})$ \\
& $(\log \mathbf{s}, \mathbf{t}) = \neuralNet(\mathbf{x}_b)$  &   &   $(\log \mathbf{s}, \mathbf{t}, \mathbf{\pi}, \mathbf{\mu}, \log \mathbf{\hat{\mathbf{s}}}) = \neuralNet(\mathbf{x}_b)$ \\ 
 & $\mathbf{y}_a = \exp(\log \mathbf{s}) \odot \mathbf{x}_a + \mathbf{t}$  &   & $\mathbf{y}_a =  \sigma^{-1} (f(\mathbf{x}_a, \mathbf{\pi}, \mathbf{\mu}, \log \mathbf{\hat{\mathbf{s}}})) \odot \exp(\log \mathbf{s})  + \mathbf{t}$ \\
 & $\mathbf{y} = (\mathbf{y}_a, \mathbf{x}_b)$  &   & $\mathbf{y} = (\mathbf{y}_a, \mathbf{x}_b)$ \\

  \hline
  
\multirow{4}{*} {\shortstack[l]{Conditional Affine \\ Coupling}} &
  $\mathbf{x}_a, \mathbf{x}_b = \osplit(\mathbf{x})$  & \multirow{4}{*} {\shortstack[l]{Conditional Affine  \\ Injector}}  &  \\
 & $(\log \mathbf{s}, \mathbf{t}) = \neuralNet(\mathbf{x}_b, \mathbf{c})$ &  & $(\log \mathbf{s}, \mathbf{t}) = \neuralNet(\mathbf{c})$  \\
  & $\mathbf{y}_a = \exp(\log \mathbf{s}) \odot \mathbf{x}_a + \mathbf{t}$ & & $\mathbf{y} = \exp(\log \mathbf{s}) \odot \mathbf{x} + \mathbf{t}$ \\
 &  $\mathbf{y} = (\mathbf{y}_a, \mathbf{x}_b)$ & & \\

 \hline
    \end{tabular}
      \caption{\small \sflow{} layers for either  unconditional \cite{mahajan2020normalizing} or conditional \cite{lugmayr2020srflow} flow models used as our backbones. Here $\mathbf{x},\mathbf{c}, \mathbf{y}$ indicate input, condition and output respectively. $\osplit$, $\neuralNet$ denote the split operation and the regular neural flow network operations. $\mathbf{x}_a, \mathbf{x}_b$ are the two splits, and $\log \mathbf{s}, \mathbf{t}, \mathbf{\pi}, \mathbf{\mu}, \log \mathbf{\hat{\mathbf{s}}}$ are transformation parameters for $\mathbf{x}_a$ produced by the network function $\neuralNet$ acting on $\mathbf{x}_b$. For mixture affine coupling, $f(\mathbf{x}_a, \mathbf{\pi}, \mathbf{\mu}, \log \mathbf{\hat{\mathbf{s}}}):=\sum_i \pi_i \sigma((\mathbf{x}_a-\mathbf{\mu}_i) \odot  \exp(-\log \mathbf{\hat{s}_i}))$, and $ \sigma(\cdot)$ indicates the sigmoid function \cite{ho2019flow++}.}
  \label{tab:flow_step}
  \vspace{-0.3cm}
\end{table*}

Lately, several \textit{unconditional generative flow} models have emerged that extends the early flow models to multi-scale architectures with split couplings that allow for efficient inference and sampling \cite{kingma2018glow,chen2019residual,ho2019flow++,mahajan2020normalizing}. For instance, \cite{kingma2018glow} introduces invertible $1 \times 1$ convolutions to encode non-linearity in the data distribution for the unconditional setup. \cite{hoogeboom2019emerging} introduces more general $d \times d$ invertible convolutions to enlarge the receptive field. \cite{chen2019residual} exploits residual blocks of flow layers (\ie, a flexible family of transformations) where only Lipschitz conditions are used for enforcing invertibility. \cite{ho2019flow++} improves the coupling layer with variational dequantization, continuous mixture cumulative distribution function, and self-attention. The self-attention is applied directly to the intrinsic neural function of the coupling layer. Because of the nature of the affine coupling layer, the attention is not required to be invertible. Besides, this direct attention application merely learns the dependencies within one of the two splits of channel-wise flow dimensions, and thus its receptive field is greatly limited. In contrast, our introduced attentions are independent flow layers that are invertible and can learn more general and better range dependencies across different splits of flow feature maps\footnote{For the similar purpose, a concurrent work \cite{zha2021invertible} also introduces invertible attentions. The major difference is that  \cite{zha2021invertible} employs Lipschitz constraints over the modules for the invertibility, which is similar to the technique presented in \cite{chen2019residual}. However, the Lipschitz constraints are generally hard to satisfy, leading to inferior results using \cite{zha2021invertible} for invertible models.}. In other words, \cite{ho2019flow++} models within-split dependencies while ours learns cross-split correlations, and hence both are complementary for each other. More recently, \cite{mahajan2020normalizing} models channel-wise dependencies through multi-scale autoregressive priors. The introduced dependency modeling is limited on latent space, and hence it can be complementary to our exploited attentions on intermediate flow dimensions.

Likewise, various \textit{conditional flow models} have appeared aiming at conditional image synthesis \cite{sun2019dual,lu2020structured,pumarola2020c,sorkhei2020full,lugmayr2020srflow}. For instance, \cite{sun2019dual} exploits two invertible networks for source and target and a relation network that maps the latent spaces to each other. In this way, conditioning information can be leveraged at the appropriate hierarchy level and hence, can overcome the restriction of using raw images as input. Similarly, \cite{pumarola2020c} exploits a parallel sequence of invertible mappings in which a source flow guides the target flow at every step. \cite{sorkhei2020full} introduces conditioning networks that allow all operations in the target-domain flow conditioned on the source-domain information. For better conditioning, \cite{lugmayr2020srflow} exploits conditional affine coupling layers that accept the source domain’s feature maps extracted by one external neural network as the conditions. 
To our knowledge, these conditional flow models rarely learn suitable range dependencies in deep normalizing flow networks.

\parsection{Attention Models}
To address the problem of missing global information in convolutional operations, attention mechanisms have emerged. They can better model deep network layer interactions \cite{vaswani2017attention,wang2018non,xu2018attngan,zhang2019self,bello2019attention,wang2020attentionnas,dosovitskiy2020image,liu2021swin}. In particular, self-attention calculates the response at a position in a sequence by attending to all positions within the same sequence allowing for long-range interactions without an increase in the number of parameters. For instance, \cite{parkbam2018bam, woo2018cbam, wang2020attentionnas} introduce map-based attention to improve the performance of convolutional networks on image recognition, where spatial attention maps are learned to scale the features given by convolutional layers. \cite{vaswani2017attention} integrates the scaled dot-product attention with its multi-head versions to construct the state-of-the-art attention (i.e., \emph{transformer}), which has become a de-facto standard for natural language processing tasks. \cite{dosovitskiy2020image,liu2021swin} achieve the state-of-the-art in a broad range of vision tasks by further applying the vanilla transformer to sequences of image patches. \cite{chen2018attention,ma2018gan,zhang2019self,cheng2020sequential,jiang2021transgan,hudson2021generative} exploit conventional attentions or transformer-based attentions in the context of other generative models like GANs to capture long-range dependencies for better image generation. Despite such remarkable progress, attention models have rarely been explored for flow-based generative models, where each neural operation is constrained to preserve tractability of the inverse and Jacobian determinant computation. To fill the gap, our proposed invertible attentions provide valuable solutions that enable such regular attentions, e.g, \emph{map-based attention} and \emph{transformer-based attention}, to work well in the context of generative flows.
\section{Overview and Background}
\label{sec3}
This paper introduces two invertible attention mechanisms to better model the network's depth dependencies for unconditional and conditional flow-based generative models\footnote{Our paper focuses on studying invertible flows that allow for both efficient exact inference and sampling.}. Our modeling is capable of producing more efficient flow models. Below we provide an overview of the unconditional and conditional generative flow models, followed by an outline of the proposed attention mechanisms. 

\smallskip
\noindent
\textit{Unconditional flow}: In this setup, the generative flows aim at learning invertible transformations (\ie, $f_{\theta},g_{\theta}$, where $\mathbf{z} = f_{\theta}(\mathbf{x})=g_{\theta}^{-1}(\mathbf{x})$ with model parameters $\theta$) between a simple distribution $\mathbf{z}\sim p_{\theta}(\mathbf{z})$ and a complex one $\mathbf{x} \sim p_{\theta}(\mathbf{x})$. The function $f_{\theta}$ (and, likewise, $g_{\theta}$) are parameterized by an invertible neural network, consisting of a sequence of $L$ invertible functions $f_{\theta_i}$. Hence, the network model is typically called as a (normalizing) flow: $f_{\theta}=f_{\theta_1}\circ f_{\theta_2}\circ \ldots f_{\theta_L}$, mapping the simple distribution density on the latent variable $\mathbf{z}$ to the complex distribution density on the data $\mathbf{x}$:
\begin{equation}
\begin{aligned}
\mathbf{x} \overset{f_{\theta_1}}{\longleftrightarrow} \mathbf{h}_{1} \overset{f_{\theta_2}}{\longleftrightarrow} \mathbf{h}_{2} \cdots \overset{f_{\theta_L}}{\longleftrightarrow} \mathbf{z}.
\label{eq:unflowtrans}
\end{aligned}
\end{equation}

Given the log-likelihood of $p_{\theta}(\mathbf{z})$, the change of variables formula enables us to compute the log-likelihood of the data $\mathbf{x}$ under the transformation $f_{\theta}$:
\begin{equation}
\vspace{-0.2cm}
\begin{aligned}
\log p_{\theta}(\mathbf{x}) & =
\log p_{\theta}(\mathbf{z}) + \log \lvert \det (\nicefrac{\partial \mathbf{z}}{\partial \mathbf{x}}) \lvert \\ & = \log p_{\theta}(f_{\theta}(\mathbf{x})) + \sum\limits_{i=1}^{L} \log \;\lvert \det (\nicefrac{\partial \mathbf{h}_{i}}{\partial \mathbf{h}_{i-1}}) \lvert,
\label{eq:unconflow}
\end{aligned}
\end{equation}
where $\nicefrac{\partial \mathbf{h}_{i}}{\partial \mathbf{h}_{i-1}}$ is the Jacobian of the invertible transformation $f_{\theta_i}$ moving from $\mathbf{h}_{i-1}$ to $\mathbf{h}_{i}$ with $\mathbf{h}_0\equiv\mathbf{x}$. The scalar value $\log \;\lvert \det J_{\theta_i} \lvert$ is the log-determinant of the Jacobian matrix\footnote{Flow-based generative models choose transformations whose Jacobian is a triangular matrix for tractable computation of log-det.}. The likelihood of $p_{\theta}(\mathbf{z})$ is commonly modeled as Gaussian likelihood, \textit{e.g.}, $p(\mathbf{z}) = \mathcal{N}(\mathbf{z}\,|\,\mu,\sigma)$. The exact likelihood computation allows us to train the network by minimizing the negative log-likelihood (NLL) loss.

\smallskip
\noindent
\textit{Conditional flow:} In this setting, the invertible network $f_{\theta}$ maps the input data-condition pair $(\mathbf{x}, \mathbf{c})$ to a latent variable $\mathbf{z}=f_{\theta}(\mathbf{x};\mathbf{c})$. Here, the data $\mathbf{x}$ is reconstructed from the latent encoding $\mathbf{z}$ conditioning on $\mathbf{c}$ as $\mathbf{x}=f_{\theta}^{-1}(\mathbf{z};\mathbf{c})$. The log-likelihood of the data $\mathbf{x}$ is computed as
\begin{equation}
\begin{aligned}
\vspace{-0.2cm}
\log p_{\theta}(\mathbf{x} | \mathbf{c}) & = \log p_{\theta}(f_{\theta}(\mathbf{x}; \mathbf{c})) + \sum\limits_{i=1}^{L} \log \;\lvert \det (\nicefrac{\partial \mathbf{h}_{i}}{\partial \mathbf{h}_{i-1}}) \lvert,
\label{eq:conflow}
\end{aligned}
\vspace{-0.1cm}
\end{equation}
\noindent
where, $\mathbf{h}_{i} = f_{\theta_i}(\mathbf{x};\mathbf{c})$. 
For both the unconditional and conditional flow models, the design of flow layers generally respects the protocol that computing the inverse and Jacobian determinant of the involved transformations $f_{\theta_i}$ should be tractable. In this paper, we mainly use \cite{mahajan2020normalizing} and \cite{lugmayr2020srflow} as our backbones for unconditional and conditional flow models respectively. In these backbones, the flow network is organized into $L$ flow-levels, each operating at a resolution containing $K$ number of flow-steps.
In general, each flow-level  $f_{\theta_i}$ is composed of \squeeze{}, \sflow{}, and \osplit{} operations. 
\squeeze{} trades off spatial resolution for channel dimension.
\sflow{} is commonly a series of affine coupling layers, invertible $1 \times 1$ convolutions and normalization layers. \osplit{} divides an intermediate layer $h_i$ into two halves, one of which is transformed and the other of which is left unchanged. Table~(\ref{tab:flow_step}) summarizes the functions of the main layers of \sflow{}.

To explicitly model the long-range dependencies for efficient flow models, we study two types of invertible attention mechanisms:  \textit{(i) invertible map-based (iMap) attention}: It aims at learning a weighting factor for each position in the attention dimension and scales the flow feature maps with the learned attention weights. The attention models the importance of each position in the attention dimension of flow feature maps explicitly. 
\textit{(ii) invertible transformer-based (iTrans) attention}: 
It computes the representation response at a position as a weighted sum of features of all the positions along the attention dimension. The attention weights are computed by scaled dot-product between features of all the positions. Compared to the iMap attention, it explicitly models the second-order dependencies among the distant positions along the attention dimension.

\section{Proposed Attention Flow}

\begin{figure}[t]
% \vspace{-0.2cm}
\centering
\includegraphics[width=0.4\textwidth]{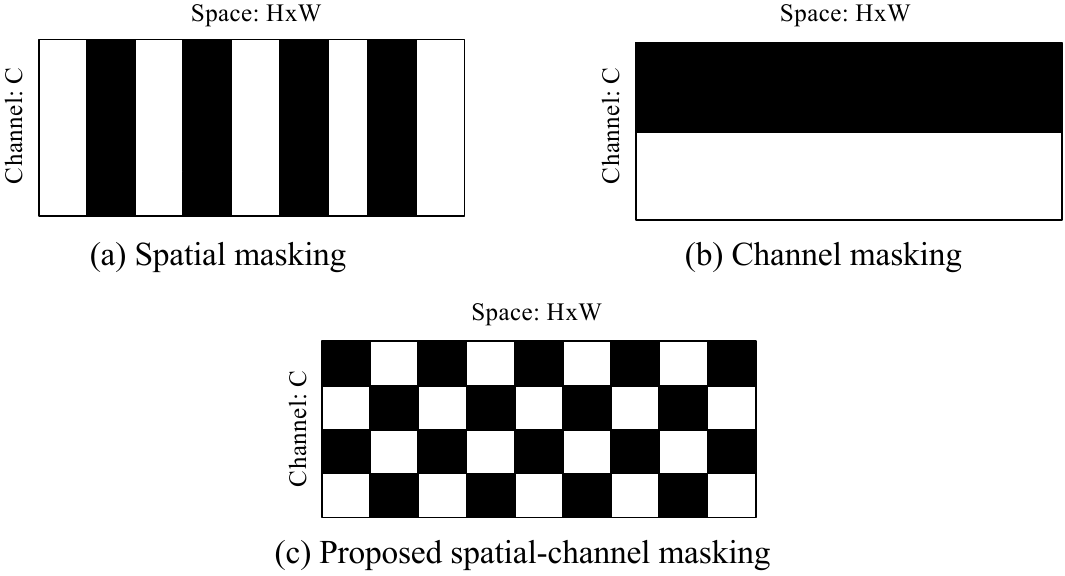}
\vspace{0.2cm}
\caption{(a) Spatial checkerboard masking \cite{dinh2017density}, (b) channel-wise masking \cite{dinh2017density,kingma2018glow}, and (c) proposed spatial-channel checkerboard masking, for the binary pattern generation on the space and channel dimensions.}
\vspace{-0.4cm}
\label{fig:checkerboards}
\end{figure}

The proposed attention flow (AttnFlow) aims at inserting invertible map-based (iMap) or transformer-based (iTrans) attention flow layers to conventional flow-based generative models (see Fig.(\ref{fig:concept}) (c)), so that the
attention learning can enhance their representation learning efficiency.
Like conventional attention mechanisms, an attention operation accepts a feature map $\mathbf{h}_{\text{in}}$ of shape $(H, W, C_{\text{in}})$ as input, and outputs an attended featured map $\mathbf{h}_{\text{out}}$ of shape $(H, W, C_{\text{out}})$ with a transformation $\mathbf{h}_{\text{out}} = G(\mathbf{h}_{\text{in}})$. In practice, attention learning consists of three steps: \textit{(i)} reshaping the input feature map $\mathbf{h}_{\text{in}}$, \textit{(ii)} computing the attention weights $\mathbf{W}_{\text{attn}}$, and \textit{(iii)} applying the learned attention weights to output $\mathbf{h}_{\text{out}}$. To integrate the introduced attention modules into generative flows, we must ensure the attention transformation $G$ preserves the tractability of inverse and Jacobian determinant computation. Hence, we introduce a checkerboard masking scheme for globally permuted binary patterns (\ie, two-split generation $\bm{x}_1, \bm{x}_2$) of the entire flow feature maps. Inspired by the existing split techniques \cite{dinh2017density, kingma2018glow}, we proposed a spatial-channel checkerboard masking scheme. As illustrated in Fig.(\ref{fig:checkerboards}) (c), 
the permutation is performed on the whole space and channel dimensions of the input feature maps. Compared to the existing spatial checkerboard masking \cite{dinh2017density} (Fig.(\ref{fig:checkerboards}) (a)) and channel-wise masking \cite{dinh2017density,kingma2018glow}  (Fig.(\ref{fig:checkerboards}) (b)) that are directly applied to generate binary patterns on the space and channel domains, the introduced spatial-channel checkerboard masking (Fig.(\ref{fig:checkerboards}) (c)) can produce more globally permuted binary patterns. As our methods learn attentions across the splits, the more permuted and staggered binary patterns allow for more complete long-range interactions.

\begin{figure*}[t]
\begin{center}
   \includegraphics[width=0.98\linewidth]{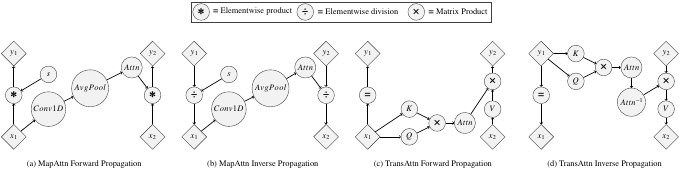}
\end{center}
\vspace{-0.2cm}
   \caption{\small Computational graph of forward and inverse propagation of the proposed Map-based (a)(b) and Transformer-based (c)(d) attention mechanisms.
   Due to the
   simple nature of the introduced split-based strategy, the involved attentions are both easily invertible and possesses a tractable Jacobian determinant.
   In (a,b), $s$ is a learnable scale parameter, and the average pooling is performed along channels (/spaces) for  spatial (/channel) attention learning. In (c,d), $K, Q, V$ are the three basic elements for the Transformer-based attention. They are computed by regular $1 \times 1$ 2D convolutions within the masking scheme, which allows for invertible operations.}
\label{fig:attnflow_main}
\vspace{-0.4cm}
\end{figure*}

The nature of the introduced global masking strategy better ensures the involved attentions can be invertible directly. Furthermore, it enables an attention transfer from one split $\bm{x_1}$ to the other split $\bm{x_2}$, which encourages interaction between the two splits along one attention dimension such as spatial dimension and channel dimension. As illustrated in Fig.(\ref{fig:attnflow_main}) (a)-(c), the overall masked flow attention operations can be roughly formulated as
\begin{subequations}
\vspace{-0.1cm}
\begin{align}
\bm{y}_1 & = \bm{x}_1 \odot \bm{s}, \label{eq:attn_forward1}\\
  \bm{y}_1 & = \bm{x}_1, \label{eq:attn_forward2} \\
 \bm{y}_2 & = \bm{x}_2 \odot f(\bm{x}_1), \label{eq:attn_forward3}
\end{align}
\label{eq:attn_forward}%
\end{subequations} 
% \end{equation*}
where Eq.\eqref{eq:attn_forward1} and Eq.\eqref{eq:attn_forward2} are for iMap and iTrans respectively, and Eq.\eqref{eq:attn_forward3} is for both. $\odot$ represents the element-wise/matrix multiplication for the proposed iMap/iTrans, and $f(\bm{x}_1)$ indicates the attention weight computation for iMap/iTrans\footnote{The new masking computation and the attention-oriented transformation make our attention operations distinct from exiting coupling layers \cite{dinh2017density,kingma2018glow} and its associated attentions like the one in \cite{ho2019flow++}.}. As shown in Fig.(\ref{fig:attnflow_main}) (b)-(d), our approach computes the inverse propagation directly as follows:
\begin{subequations}
\vspace{-0.1cm}
\begin{align}
\bm{x}_1 & = \bm{y}_1 \oslash \bm{s}, \label{eq:attn_back1}\\
  \bm{x}_1 & = \bm{y}_1, \label{eq:attn_back2} \\
 \bm{x}_2 & = \bm{y}_2 \oslash f(\bm{x}_1), \label{eq:attn_back3}
\end{align}
\label{eq:attn_back}%
\end{subequations} 
where Eq.\eqref{eq:attn_back1} and Eq.\eqref{eq:attn_back2} are for iMap and iTrans respectively, and Eq.\eqref{eq:attn_back3} is for both.  $\oslash$ indicates the element-wise/matrix division for iMap/iTrans, and $f(\bm{x}_1)$ denotes the computation of attention weights. Below we provide the details of the two introduced invertible attentions, and the corresponding Jacobian determinant computation.

\parsection{iMap Attention} Following \cite{wang2020attentionnas} that invents regular map-based attention (i.e., diagonal attention), we exploit an invertible map-based attention to scale the feature map with the learned attention weights that encodes the importance of individual flow dimensions along the attention dimension. The main difference is that we apply attention weights calculated on one split $\bm{x_1}$ to the other split $\bm{x_2}$, due to the invertiable design in Eq.\eqref{eq:attn_forward} and Eq.\eqref{eq:attn_back}. Concretely, we apply a sequence of analogous functions from \cite{wang2020attentionnas} to realize iMap over the spatial domain of flow feature maps. Mathematically, the attention weights can be calculated as 
\begin{equation}
\label{eq:imap_weight}
\mathbf{W}_\text{imap} = G_5\Big((1-\mathbf{M}) G_4\big( G_3(G_2( G_1( \mathbf{h}_\text{in} ) ) ) \big) + \mathbf{M}\mathbf{b} \Big),
\end{equation}
where $\mathbf{M}$ is the proposed checkerboard mask (Fig.(\ref{fig:checkerboards}) (c)), $\mathbf{b}$ is a learnable variable, $G_1(\mathbf{h}_\text{in})=\mathbf{M}\mathbf{h}_\text{in}$, $G_2$ is a 1D convolutional layer with kernel size as 1, which reduces the dimension of the feature response of each channel from $C_{\text{in}}$ to $C'$, and outputs a  feature map of shape $(H\times W, C')$. Without loss of generality, $G_3$ applies average pooling\footnote{Performing average pooling over the spatial domain learns channel attention, provided the spatial resolutions for train and validation are same.} to each channel dimensions and outputs an $(H \times W)$-dim vector for spatial attention learning. 
The operator $G_4$ is to reorganize the $(H \times W)$ attention weights into a $(H \times W) \times (H \times W)$ matrix, where the attention weights of shape $(H \times W)$ are placed on the diagonal of the matrix. The derived attention weight matrix $\mathbf{W}_\text{imap}$ is a diagonal matrix. The function $G_5$ corresponds to standard activation functions such as softmax and sigmoid.
Finally, we apply the attention weight matrix $\mathbf{W}_\text{imap}$ to the input feature map through matrix multiplication to obtain the attended feature map $\mathbf{h}_\text{out} = \mathbf{W}_\text{imap}\mathbf{h}_\text{in}$.
The forward and inverse propagation of \emph{AttenFlow-iMap} module are illustrated in Fig.(\ref{fig:attnflow_main}) (a)-(b). 
The Jacobian determinant of the introduced iMap transformation is computed as follows:
\begin{equation} 
\footnotesize
\begin{split}
\det(\frac{\partial \mathbf{h}_{\text{in}}}{\partial \mathbf{h}_{\text{out}}}) = \det(W_{\text{imap}})=\left(\prod_{\bm{M}_{j,:}=\mathbf{1}} G_5(\mathbf{b_j})\right)* (G_5(G^{'}(\mathbf{h}_\text{in})))^{C_{\text{in}}/2}, 
\end{split}
\label{eq:imap_jocob}
\end{equation}
\noindent
where $\bm{M}$ is the enforced mask, $C_{\text{in}}$ is the channel number of $h_{\text{in}}$, $G_5$ indicates the corresponding activation function, $G^{'}(\mathbf{h}_\text{in})=G_3(G_2(G_1(\mathbf{h}_\text{in})))$, $G_1, G_2, G_3$ are masking, 1D convolution, and average pooling respectively.

\parsection{iTrans Attention} The conventional transformer-based attention was proposed in \cite{vaswani2017attention}. The success of this type of attention mechanism mainly stems from the effective learning of second-order correlations among involved feature maps and the exploitation of three different representations for attention learning. The attention function is expressed as mapping a query $\mathbf{q}_{\text{in}}$ and a set of key-value $(\mathbf{k}_{\text{in}},\mathbf{v}_{\text{in}})$ pairs to an output $\mathbf{h}_{\text{out}}$. The query and the key are employed to learn the second-order attention weights through a scaled dot-product computation, which is further applied to the input value for the final attended output.

To introduce the transformer-based attention to flow models, as shown in Fig.(\ref{fig:attnflow_main})(c), we apply two invertible $1 \times 1$ 2D convolutions to the input feature maps to obtain a query-key pair $(\mathbf{q}_{\text{in}},\mathbf{k}_{\text{in}})$, and use the input feature maps to play the role of the value $\mathbf{v}_{\text{in}}$. The attention is applied between patches of the input following \cite{dosovitskiy2020image}. In particular, the whole input is split into $N$ patches and the iTrans attention is applied to the image patches. The primary goal is to capture the inter-patch interaction with the attention weights. In practice, we compute the attention function on a set of queries simultaneously, packed together into a matrix $\mathbf{Q}$. The keys and values are also packed together into matrices $\mathbf{K}$ and $\mathbf{V}$. The mapping process is formulated as follows:
\begin{equation}
   \mathbf{h}_\text{out} = \mathbf{W}_\text{itrans} \mathbf{V} = G_4(\frac{\mathbf{Q}\mathbf{K}^T}{\sqrt{d}})\mathbf{V},
   \label{eq:sdp}
\end{equation}
where $\mathbf{Q}=G_2(G_1(\mathbf{h}_\text{in})), ~\mathbf{K}=G_3(G_1(\mathbf{h}_\text{in})), \mathbf{V}= G_1(\mathbf{h}_\text{in})=\mathbf{M}\mathbf{h}_\text{in}$, $\mathbf{M}$ is the suggested checkerboard mask (Fig.(\ref{fig:checkerboards}) (c)),  $G_2, G_3$ correspond to two regular $1 \times 1$ 2D convolutions,  $G_4$ corresponds to the activation function. $G_2, G_3$ are computed within the introduced masking, which allows for invertible operations.
In general, dot-product values often get large to influence the final negative log-likelihood scales. Hence, inspired by \cite{vaswani2017attention}, we apply $d$ to scale the dot-product values. To achieve a general scale, we made $d$ learnable. In addition, we follow the vanilla transformer \cite{vaswani2017attention} to exploit multiple branches of scaled dot-products (Eq.\eqref{eq:sdp}) for the multi-head attention. Fig.(\ref{fig:attnflow_main})(d) show the inverse propagation of \emph{AttnFlow-iTrans}, which can be computed in a straightforward manner.

The Jacobian of the iTrans transformation, with the attended feature map being $\bm{h}_\text{out}=\bm{W}_{\text{itrans}} \bm{M} \bm{h}_{\text{in}}$, is a lower block triangular matrix, with the attention weights $\mathbf{W}_\text{itrans}$ forming the (repeated) block diagonal entries. As the determinant of a lower block triangular matrix is simply the product of determinants of the matrices along the diagonal, the Jacobian determinant of iTrans can be computed as 
\begin{equation}
\det(\frac{\partial \mathbf{h}_{\text{in}}}{\partial \mathbf{h}_{\text{out}}}) =  (\det(\mathbf{W}_\text{itrans}))^{P/2}=(\det(G_4(\frac{\mathbf{Q}\mathbf{K}^T}{\sqrt{d}})))^{P/2},
\end{equation}
where $G_4,
\bm{Q},\bm{K}, d$ are defined around Eq.\eqref{eq:sdp}, and $P$ is the patch size, \ie, feature dimension within each patch.

\section{Experimental Evaluation}

We evaluated the proposed unconditional and conditional attention flow (AttnFlow, cAttnFlow)\footnote{AttnFlow's official code: \url{https://github.com/rheasukthanker/AttnFlow}} models for image generation, image super-resolution and general image translation tasks respectively\footnote{Following \cite{mahajan2020normalizing,lugmayr2020srflow,sorkhei2020full}, we evaluate the proposed method and all the competing methods with one single run on the employed datasets.}. 
Besides, we present more experimental details and evaluations in the suppl. material.

\begin{figure*}[t]
\centering
\scriptsize
\resizebox{1\linewidth}{!}{%
\begin{tabular}{cccc}
\includegraphics[width=0.239\linewidth]{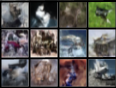} &  \includegraphics[width=0.242\linewidth]{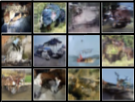} &
\includegraphics[width=0.242\linewidth]{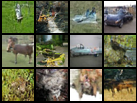} &
\includegraphics[width=0.239\linewidth]{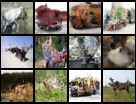}\\
(a) Flow++ (3.08 bits/dim) & (b) mARFlow  (3.24 bits/dim, 41.9 FID) & (c) \textit{AttnFlow-iMap} (3.216 bits/dim, 33.6 FID) & (d) \textit{AttnFlow-iTrans} (3.217 bits/dim, 33.8 FID) \\
\end{tabular}
}
\caption{\small Comparison of samples from the proposed models (AttnFlow-iMap, AttnFlow-iTrans) with state-of-the-art models  on CIFAR10.}
\label{fig:cifar10_samp}
\vspace{-0.3cm}
\end{figure*}

\noindent
\textbf{Image Generation.} We use two datsets \ie, MNIST \cite{lecun1998gradient} and CIFAR10 \cite{krizhevsky2009learning} for unconditional image generation.

\begin{table}[t]
\scriptsize
  \centering
   \resizebox{0.46\textwidth}{!}
    {
    %\footnotesize
    \begin{tabular}{l@{~~~}c@{~~}c@{~~}c@{~~} c@{~~} c@{~~}}
	\toprule
	Method & Levels & Steps & Channels & Parameters (MB) & {bits/dim $(\downarrow)$} \\
	\midrule
	Glow & 3 & 32 & 512 & {--} & 1.05 \\ 
	Residual Flow  & 3 & 16 & {--}  & {--} & 0.97 \\
	mARFlow & 3 &  4 & 96 & 46.01 &  0.56 (0.88$^*$) \\
	\midrule
    \textit{AttnFlow-iMap} & 3 & 4  & 96 & 46.03  &  0.43\\ 
    \textit{AttnFlow-iTrans} & 3 & 4 & 96 & 46.25 &  0.44  \\ 
    \midrule
    \textit{AttnFlow-iMap} & 3 & 2  & 96 & 23.78  &  0.41 \\ 
    \textit{AttnFlow-iTrans} & 3 & 2 & 96 & 23.89   & 0.42\\ 
    \midrule
     \textit{AttnFlow-iMap} & 3 & 2 & 48 & 8.94  &  \textbf{0.39} \\ 
    \textit{AttnFlow-iTrans} & 3 & 2 & 48 & 9.05 &  \underline{0.40} \\ 
	\bottomrule
   \end{tabular} 
   }
     \caption{\small Evaluation of sample quality on MNIST. $*$ indicates the result reported in the mARFlow paper \cite{mahajan2020normalizing}. As MNIST is a small dataset and very complex model is not at all required, the performance gets decreased when our model's complexity increases. (\textbf{Bold}: best, \underline{Underline}: second best)}
  \label{tab:mnist}
  \vspace{-0.5cm}
\end{table}

\noindent
\textit{1) AttnFlow Setup:}
The proposed AttnFlows can be applied to any off-the-shelf unconditional generative flows. For the image generation task, we utilize the architecture of mARFlow\footnote{mARFlow's official code: \url{https://github.com/visinf/mar-scf/}} \cite{mahajan2020normalizing} as the backbone of AttnFlows, where our proposed iMap and iTrans attention flow layers can be inserted.
Each level of mARFlow sequentially stacks an actnorm layer, an invertible $1 \times 1$ convolution layer, and a coupling layer. 
Under the mARFlow backbone, we inserted our proposed attention modules (either iMap or iTrans) into one of the following four positions: \textit{(i)} Before actnorm (pos-1), \textit{(ii)} after actnorm (pos-2), \textit{(iii)} after invertible convolution (pos-3), and \textit{(iv)} after coupling (pos-4).
To study AttnFlows' efficiency, we evaluate their various setups on the numbers of flow-levels, flow-steps, and channels.
We use sigmoid for the activation function, and empirically set the patch number as $N=4$ for AttnFlow-iTrans.

\noindent
\textit{2) Competing Methods:} We compare four state-of-the-art unconditional generative flows, i.e., Glow \cite{kingma2018glow}, Flow++ \cite{ho2019flow++}, Residual Flow \cite{chen2019residual} and mARFlow \cite{mahajan2020normalizing}. Our AttnFlows' architecture is based on mARFlow with coupling layers closest to Glow. By comparison, Flow++ does not include the \osplit{} operation, and uses a different uniform dequantization. Hence, the comparison with Glow and mARFlow serves as a better ablation to measure the effectiveness and efficiency of AttnFlows. Besides, we compare the concurrent work \cite{zha2021invertible}, with its two variants (iResNet-iDP, iResNet-iCon), which applies Lipschitz constraints to dot-product and concatenation attentions under the specific flow framework (iResNet) \cite{chen2019residual}.
For a reference, we also compare one representative GAN model, \ie, DCGAN \cite{radford2015unsupervised}.

\noindent\textit{3) Comparison:} 
Table (\ref{tab:mnist}) and Table (\ref{tab:cifar}) summarize the quantitative results of our AttnFlows and the competing methods on MNIST and CIFAR10.
For evaluation, we use per-pixel log-likelihood metric in bits/dims. Further, we use three more standard metrics, \ie, Fr\'echet Inception Distance (FID) \cite{heusel2017fid}, inception scores \cite{salimans2016improved} and Kernel Inception Distance (KID) \cite{binkowski2018demystifying}, to measure the generated sample quality on CIFAR10. From the results, we can see that both of our AttnFlow-iMap and AttnFlow-iTrans clearly outperform the backbone mARFlow with similar model complexities (\ie, the same level and step numbers), and our AttnFlows can achieve better results than the other state-of-the-art flow models\footnote{ Flow++\cite{ho2019flow++} merely reported its performance on CIFAR10. After transferring its implementation from CIFAR10 to MNIST, its bits/dim is 0.66 that is also clear worse than ours
(0.39).}. Furthermore, our lighter model (with a smaller number of steps or a smaller channel) typically achieve comparable performances (or even better results) compared to those heavier mARFlow models. In particular, the proposed method achieves remarkable improvement (i.e., 0.17 bits/dim) over mARFlow with about 5$\times$ smaller parameter size and 2$\times$ less steps/channels (Table (\ref{tab:mnist})). For CIFAR10, our models (Channel=256) get visibly better FIDs, Incepts and KIDs over mARFlow (Table (\ref{tab:cifar})). The visual comparison in Fig.(\ref{fig:cifar10_samp}) shows that the proposed models have much clear visual quality compared to the competing methods. Despite the intuitive superiority of the proposed spatial-channel masking against the existing ones \cite{dinh2017density,kingma2018glow} (Fig.(\ref{fig:checkerboards})), we evaluate these maskings and the random binary masking with our AttnFlow-iMap on MNIST. The bits/dim are 0.99 (Spatial), 0.75 (Channel), 0.50 (Random), 0.39 (Ours), showing ours’ clear advantage.

\begin{table}[]
  \centering
  \footnotesize
    
    \resizebox{0.48\textwidth}{!}
    {
    \begin{tabular}{l@{~~~}c@{~~}c@{~~}c@{~~}c@{~~}c@{~~}c@{~~}c@{~~}c@{~~}}
	\toprule
	Method & Level & Step & Channel &  Parameter (MB) & {bits/dim $(\downarrow)$} & {FID $(\downarrow)$} & {Incep $(\uparrow)$} & {KID $(\downarrow)$}\\
	\midrule
	DCGAN  & {--} & {--} & {--} & {--} & {--} & 37.1 & 6.4 & {--}\\
	\midrule
	Glow  & 3 & 32 & 512 & {--} & 3.35 & 46.9 & {--} & {--} \\ 
	Flow++  & 3 &  {--} & 96 & {--} & 3.29 & 46.9 & {--} & {--} \\ 
	Residual Flow  & 3 & 16 &  {--} & {--} & 3.28 & 46.3 & 5.2 & {--}\\
	iResNet-iDP & {--} & {--} & {--} & {--} & 3.65 & {--} & {--} & {--}\\ 
	iResNet-iCon & {--} & {--} & {--} & {--} & 3.39 & {--} & {--} & {--}\\ 
    mARFlow  & 3 & 4 & 96 & 46.01 & 3.27 (3.254*) & (40.5*) & (5.8*) & (0.033*) \\
	mARFlow  & 3 & 4 & 256 &  252.77 & 3.24 (3.222$^*$) & 41.9 (33.9$^*$) &  5.7 (6.5$^*$) & (0.026$^*$) \\
	\midrule
    \textit{AttnFlow-iMap} & 3 & 4 & 96 & 46.03 & 3.247 & 40.5  & 6.0 & 0.031\\ 
    \textit{AttnFlow-iTrans} & 3 & 4 & 96 & 46.25 & 3.248 & 40.2  & 5.9 & 0.032\\ 
    \midrule
    \textit{AttnFlow-iMap} & 3 & 4 & 256 & 252.79 & \textbf{3.216} &  \textbf{33.6}  & \underline{6.6} & \textbf{0.025}\\ 
    \textit{AttnFlow-iTrans} & 3 & 4 & 256 & 253.01 & \underline{3.217} & \underline{33.8}  & \textbf{6.7} & \textbf{0.025}\\
	\bottomrule
    \end{tabular}
    }
    \caption{\small Evaluation of sample quality on CIFAR10. Note that $*$ indicates the results for the ICML workshop version of mARFlow \cite{mahajan2020normalizing}. (\textbf{Bold}: best, \underline{Underline}: second best)}\label{tab:cifar}
  \vspace{-0.4cm}
\end{table}

\begin{figure}[t]
\begin{center}
   \includegraphics[width=0.8\linewidth]{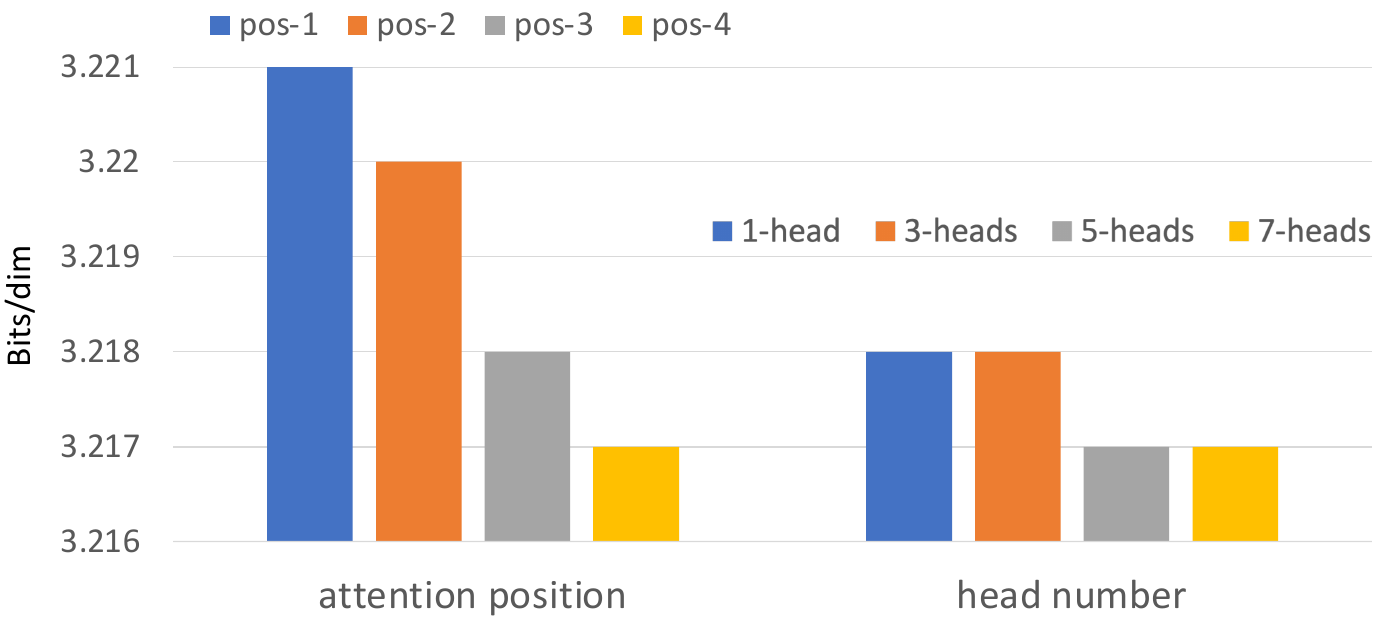}
\end{center}
\vspace{-0.2cm}
\caption{\small Ablation studies of the proposed attention on different positions in the flow layers (pos-1: before actnorm, pos-2: after actnorm,pos-3: after permutation, pos-4: after coupling layer) and different number of attention heads (1 head, 3 heads, 5 heads, 7 heads) for AttnFlow-iTrans on CIFAR10.}
\label{fig:abation}
\vspace{-0.3cm}
\end{figure}

\noindent
\textbf{Ablation Study.}
As shown in Fig.(\ref{fig:abation}), we perform the ablation test of the proposed attention models in the following two settings: \textit{(i)} different attention positions, and \textit{(ii)} different head numbers.
We observe that inserting the attention layers in the position after the permutation layer (pos-3) and after the coupling layer (pos-4) are the most favourable. On the other hand, the use of more than 5 attention heads for AttnFlow-iTrans does not provide a clear improvement.

\noindent
\textbf{Image Super-Resolution.}
We follow \cite{lugmayr2020srflow} to use CelebA dataset split for image super-resolution (SR) task \cite{liu2015deep}.

\noindent
\textit{1) cAttnFlow Setup:} Our conditional AttnFlow (cAttnFlow) is based on the architecture of the SRFlow model \cite{lugmayr2020srflow}\footnote{SRFlow's official code: \url{https://github.com/andreas128/SRFlow/}}. The flow network is organized into $L=4$ flow-levels, each of which operate a specific resolution of $\nicefrac{H}{2^l} \times \nicefrac{W}{2^l}$ ($H \times W, L$ indicate the resolution of HR images and the $l$-th flow-level respectively). Each flow-level is composed of $K$ flow-steps. Each flow-step stacks four different layers: \textit{(i)} Acnorm, \textit{(ii)} $1 \times 1$ invertible convolution, \textit{(iii)} affine injector, and \textit{(iv)} conditional affine layers. Similar to image generation, we insert our proposed attentions after the existing flow layers in each level of SRFlow.

\noindent
\textit{2) Competing Methods:} Following SRFlow, we compare our results with bicubic and other recent SR methods, which includes ESRGAN \cite{wang2018esrgan} and SRFlow \cite{lugmayr2020srflow}. As SRFlow is our cAttnFlow's backbone, comparison against it helps us realize the improvement using our introduced attentions.

\noindent
\textit{3) Comparison:} Table (\ref{tab:cleba}) reports the comparison of our cAttnFlow against the competing methods in terms of four standard metrics, including SSIM, PSNR, LR-PSNR and LPIPS. The results imply that our model can achieve the best balance among the four used metrics compared to the competing methods. The improvements of our cAttnFlow over the backbone SRFlow are visible at two different level model complexities, showing that our introduced attention can enhance the efficiency of the flow models. 
The visual comparison in Fig.(\ref{fig:celeba_sample}) shows that the outputs from our cAttnFlows are comparable or better than those from the others.
% For visual comparison refer to the supplementary material. 

\begin{table}[]
\centering%
 
   \resizebox{0.48\textwidth}{!}
    {
    \begin{tabular}{l@{~~~}c@{~~}c@{~~}c@{~~}c@{~~}c@{~~}c@{~~}c@{~~}}
    \toprule
  Method & Levels & Steps  & Parameters (MB)  & SSIM ($\uparrow$) & PSNR ($\uparrow$) & LR-PSNR ($\uparrow$)  & LPIPS ($\downarrow$)\\
    \midrule
     Bicubic & {--} & {--}  & {--}  & 0.63   & 23.15 & 35.19 & 0.58\\
     ESRGAN     & {--} & {--}  & {--}   & 0.63 & 22.88 &  34.04  & \textbf{0.12}\\
    \midrule
      SRFlow  & 1 & 1  & 6.622 &  0.67 & \textbf{25.57}  & 44.20 & 0.23\\
     SRFlow  & 2 & 8 & 13.25  &  \underline{0.73}  & 25.47 & 38.94  &  0.17\\
    \midrule
    \textit{cAttnFlow-iMap} & 1 & 1  & 6.623  & 0.71 & \underline{25.50} & \textbf{44.75}  & 0.19\\ 
    \textit{cAttnFlow-iTrans} & 1 & 1  &  6.630   & \underline{0.73} & \underline{25.50} & \underline{44.23} & 0.18 \\ 
   
    \midrule
    \textit{cAttnFlow-iMap}  & 2 & 8  &  13.30 & \textbf{0.74} & 25.38 & 41.88 & 0.17 \\ 
    \textit{cAttnFlow-iTrans}  & 2 & 8  & 13.93 & \underline{0.73} & 25.24  & 42.49 & \underline{0.16}\\ 
    \bottomrule
    \end{tabular}
    }
    \caption{\small Results for $8\times$ SR on CelebA. We report average SSIM, PSNR, LR-PSNR and LPIPS scores for SRFlow and ours at different temperatures (0.1-0.9). (\textbf{Bold}: best, \underline{Underline}: second best)}\label{tab:cleba}
    \vspace{-0.4cm}
\end{table}

\begin{figure}[h!]
    \centering%
    \newcommand{\size}{0.18}%
    \newcommand{\img}[1]{%
    \includegraphics[width=\size\linewidth]{figures/sotaCMbic8/#1/lq}~~~~%
    \includegraphics[width=\size\linewidth]{figures/sotaCMbic8/#1/ESRGANn23px2e-1.jpg}~~%
    \includegraphics[width=\size\linewidth]{figures/sotaCMbic8/#1/srflow.png}~~%
    \includegraphics[width=\size\linewidth]{figures/sotaCMbic8/#1/imap.png}~~%
    \includegraphics[width=\size\linewidth]{figures/sotaCMbic8/#1/isdp.png}~~%

    }%
    \img{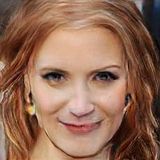}
    \img{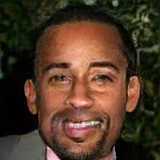}
    \img{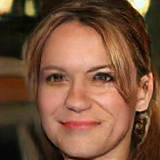}
        \vspace{-0.2cm}
    \resizebox{1\linewidth}{!}{%
    \begin{tabular}{ C{2.5cm} C{2.5cm} C{2.5cm} C{2.5cm} C{2.5cm} }
        Input  &
        ESRGAN~\cite{wang2018esrgan}  & SRFlow~\cite{lugmayr2020srflow} & \textit{cAttnFlow-iMap} &  \textit{cAttnFlow-iTrans}
    \end{tabular}}%
    \caption{\small Super-resolved samples of the proposed cAttnFlows and the state-of-the-art models for $8\times$ face SR on the CelebA dataset.}
    \label{fig:celeba_sample}
    \vspace{-0.3cm}
\end{figure}

\begin{figure}[h!]
\centering
\scriptsize
\resizebox{1.0\linewidth}{!}{%
\begin{tabular}{ccc}
% \toprule
\includegraphics[width=0.33\linewidth]{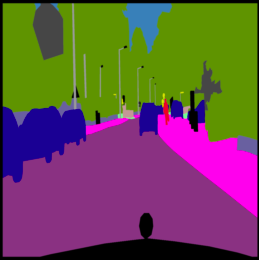} &
\includegraphics[width=0.33\linewidth]{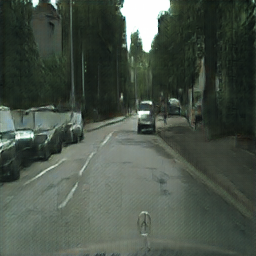} &
\includegraphics[width=0.33\linewidth]{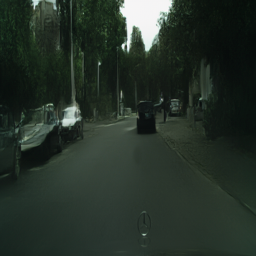}\\
(a) Input  & (b) Pix2PixGAN \cite{isola2017image} & (c) Dual-Glow \cite{sun2019dual}\\

\includegraphics[width=0.33\linewidth]{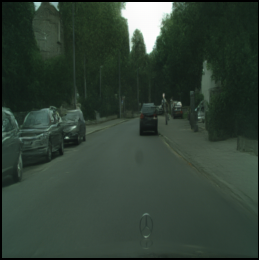} &
\includegraphics[width=0.33\linewidth]{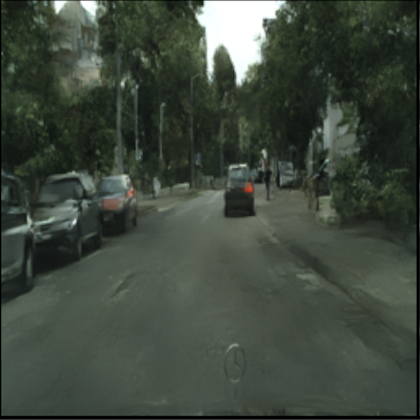} &
\includegraphics[width=0.33\linewidth]{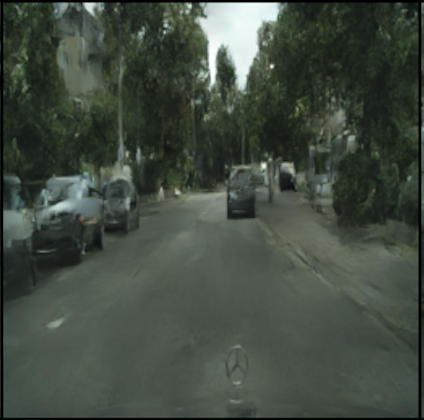}\\
(d) Full-Glow \cite{sorkhei2020full} & (e) \textit{cAttnFlow-iMap} & (f) \textit{cAttnFlow-iTrans} \\

\end{tabular}
% \vspace{0.2cm}
}
% \caption{\small 
\caption{Generated samples of the proposed cAttnFlows and the state-of-the-art models for image translation on the Cityscapes dataset. The competing methods and ours are conditioned on the semantic segmentation labels (a) to synthesize the RGB images with the resolution being of $256 \times 256$.
}
\label{fig:cityscapes}
\vspace{-0.495cm}
\end{figure}

\noindent
\textbf{Image Translation.}
We use Cityscapes  \cite{Cordts2016Cityscapes} to evaluate the proposed cAttnFlows for image translation, where segmentation label images are translated into RGB images.

\noindent
\textit{1) cAttnFlow Setup:} Our conditional AttnFlow (cAttnFlow) is based on the conditional flow (Full-Glow)\footnote{Full-Glow's official code: \url{https://github.com/MoeinSorkhei/glow2}} \cite{sorkhei2020full} model. 
The normalizing flow network is organized into $L=2$ flow-levels, each of which operate a specific resolution of $\nicefrac{H}{2^l} \times \nicefrac{W}{2^l}$, where $H \times W$ indicates the resolution of input images and the $l$-th flow-level respectively. Each flow-level is composed of $K=8$ flow-steps. Note that our flow model is much smaller than the Full-Glow model that consists of 4 levels and each contains 16 steps (i.e., $L=4, K=16$). Each flow-step stacks four different layers: 1) Acnorm, 2) $1 \times 1$ invertible convolution, 3) affine injector, and 4) conditional affine layers. As done for image generation, we also insert our proposed flow attentions after the existing flow layers in each level of the Full-Glow model.

\noindent
\textit{2) Competing Methods:} Following \cite{sorkhei2020full}, we compare with the state-of-the-art conditional flow methods, C-Glow \cite{lu2020structured}, Dual-Glow \cite{sun2019dual}, and Full-Glow \cite{sorkhei2020full}. We also compare the GAN model (Pix2Pix) \cite{isola2017image} for a refrence. As we use  Full-Glow as our cAttnFlow's backbone, we will focus on the comparison with it, which can clearly show the improvement using our introduced attention mechanisms.

\noindent
\textit{3) Comparison:} 
For likelihood-based models, we follow \cite{sorkhei2020full} to measure the conditional bits per dimension, $-\mathrm{log_2} \ p(\mathbf{x}_{b} | \mathbf{x}_{a})$, as a metric of how well the conditional distribution learned by the model matches the real conditional distribution, when tested on held-out examples.
Table (\ref{tab:cityscapes}) summarizes the results of the proposed cAttnFlows and its competitors. The comparison shows that the proposed cAttnFlows can achieve better performances than the state-of-the-art conditional flow models. In particular, compared to the backbone model (Full-Glow), the proposed cAttnFlows achieve better bits/dim with about 5$\times$ smaller parameter size and 2$\times$ less levels/steps, showing that the proposed attention can highly enhance the efficiency of flow models. The visual comparison in Fig.(\ref{fig:cityscapes}) shows that the synthesized images of our cAttnFlow are more visually pleasing (e.g, owning clearly richer texture details and better illumination) compared to the competing generative flow models, and they look relatively comparable with that produced by Pix2PixGAN \cite{isola2017image}.

\begin{table}[]
  \centering
  \footnotesize
    
    \resizebox{0.46\textwidth}{!}
    {
    \begin{tabular}{l@{~~~}c@{~~}c@{~~}c@{~~}c@{~~}}
	\toprule
	Method & Levels & Steps &  Parameters (MB) & {Conditional bits/dim $(\downarrow)$} \\
	\midrule
	C-Glow v.1  & {--} & {--} & {--} & 2.568  \\ 
	C-Glow v.2  & {--} & {--} & {--} & 2.363  \\ 
	Dual-Glow   & {--} & {--} & {--} & 2.585 \\ 
    Full-Glow & 4 & 16 & 155.33 & 2.345 \\ 
	\midrule
    \textit{cAttnFlow-iMap} & 2 & 8 & 34.68 & \textbf{2.310} \\ 
    \textit{cAttnFlow-iTrans} & 2 & 8 & 34.70 & \underline{2.314} \\ 
	\bottomrule
    \end{tabular} 
    }
    \caption{\small Quantitative results of the proposed AttnFlow and the state-of-the-art models on the Cityscapes dataset for label $\rightarrow$ photo image translation. (\textbf{Bold}: best, \underline{Underline}: second best)}\label{tab:cityscapes}
  \vspace{-0.4cm}
\end{table}

\section{Conclusion and Future Work}\label{sec:conclu}

This paper introduces invertible map-based and transformer-based attentions for both unconditional and conditional generative normalizing flows. The proposed attentions are capable of learning network dependencies efficiently to strengthen the representation power of flow-based generative models. The evaluation on image generation, super-resolution and image translation show clear improvement of our proposed attentions over the used unconditional and conditional flow-based backbones. 

As conventional attention mechanisms, one of our models' major limitations lies in its unsatisfactory scaling ability to deeper neural networks such as full SRFlow \cite{lugmayr2020srflow}, due to the common attention vanishing problem studied in \cite{dong2021attention}. As a future work, we will follow \cite{dong2021attention} to address the problem in the context of deeper invertible flow models.

% \section*{Acknowledgments}
\smallskip
\noindent\small{\textbf{Acknowledgments.}} This work was supported in part by the ETH Z\"urich Fund (OK), an Amazon AWS grant, and an Nvidia GPU grant. Suryansh Kumar’s project is supported by ``ETH Z\"urich Foundation 2019-HE-323, 2020-HS-411'' for bringing together best academic and industrial research.  This work was also supported by the Singapore Ministry of Education (MOE) Academic Research Fund (AcRF) Tier 1 grant (MSS21C002). The authors would like to thank Andreas Lugmayr for valuable discussions.

%%%%%%%%% REFERENCES
% {\small
% \bibliographystyle{ieee_fullname}
% \bibliography{egbib}
% }

\clearpage

\appendix
\normalsize
\section*{Supplementary Material}

% \begin{abstract}

The supplementary document includes: (i) further illustration of the suggested Jacobian computation, (ii) neural architecture design details of the proposed attention flows (AttnFlows), (iii) detailed experimental settings, (iv) training details and curves,  (v) more visual results on MNIST, CIFAR10, CelebA and Cityscapes,  %(E) an evaluation on a new backbone for image translation on Cityscapes, 
(vi) better/more visualization plots of the ablation study presented in the main paper on the used datasets, as well as some more results for different iTrans-based attention head numbers on CelebA, (vii) pseudo code of the proposed key components (i.e., iMap and iTrans), and (viii) further remarks on the possible future directions.

% \end{abstract}

\section{Jacobian Computation}

\begin{figure*}[h!]
\begin{center}
   \includegraphics[width=0.84\linewidth]{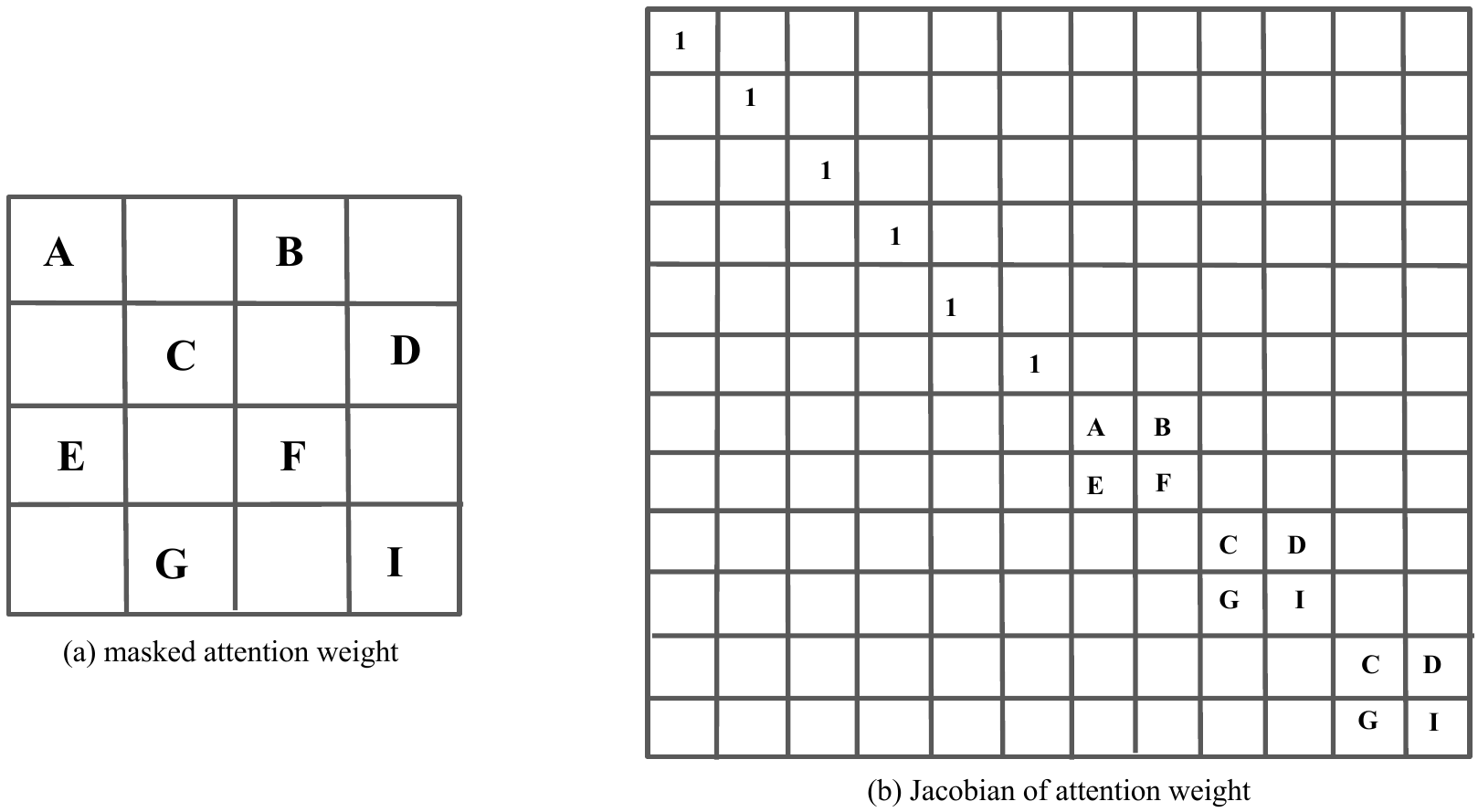} 
\end{center}
   \caption{\small (a) Example of masked attention weight. (b) Example of Jacobian matrix structure.}
\label{fig:jacobian}
\end{figure*}

The suggested masking over the learned attention weights (Eqn.6 and Eqn.8 of the main paper) additionally leads to tractable Jacobian computation, i.e., Eqn.7 and Eqn.9 of the main paper. One example of the masked attention weight is illustrated in Fig.(\ref{fig:jacobian}) (a). For this example, the resulting Jacobian is a block-lower triangular matrix, as shown in Fig.(\ref{fig:jacobian}) (b). This is because one part of the input is made to depend on the other portion of the input. In this case, the determinant of a block triangular matrix can be easily computed by the product of the determinants of its block diagonal matrices. The resulting Jocobian determinant enables us to compute the log-likelihood of  data efficiently by Eqn.2 (or Eqn.3) of the main paper.

\section{AttnFlow Network Architecture}

\begin{figure*}[h!]
\begin{center}
   \includegraphics[width=0.36\linewidth]{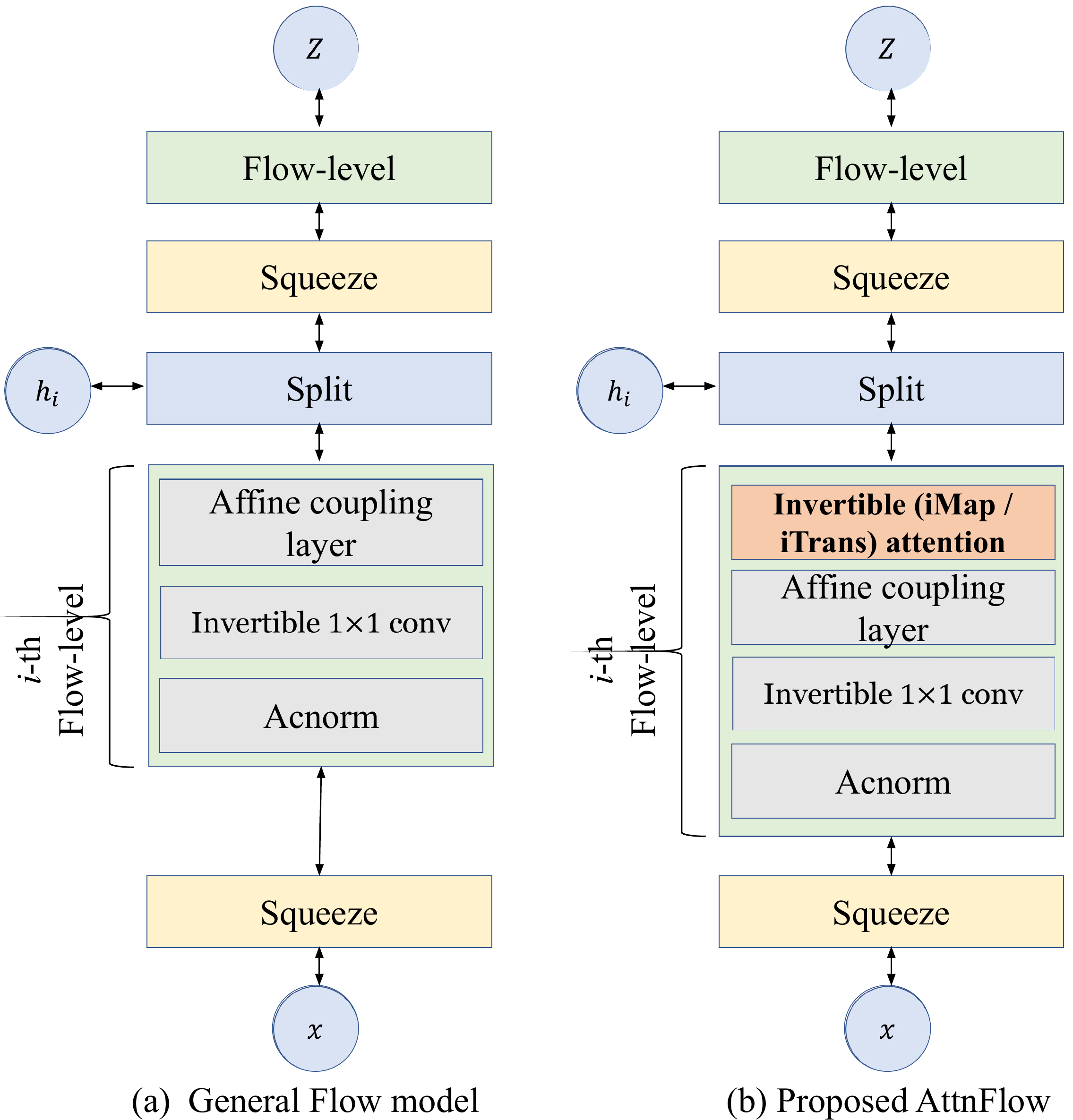} 
  \includegraphics[width=0.62\linewidth]{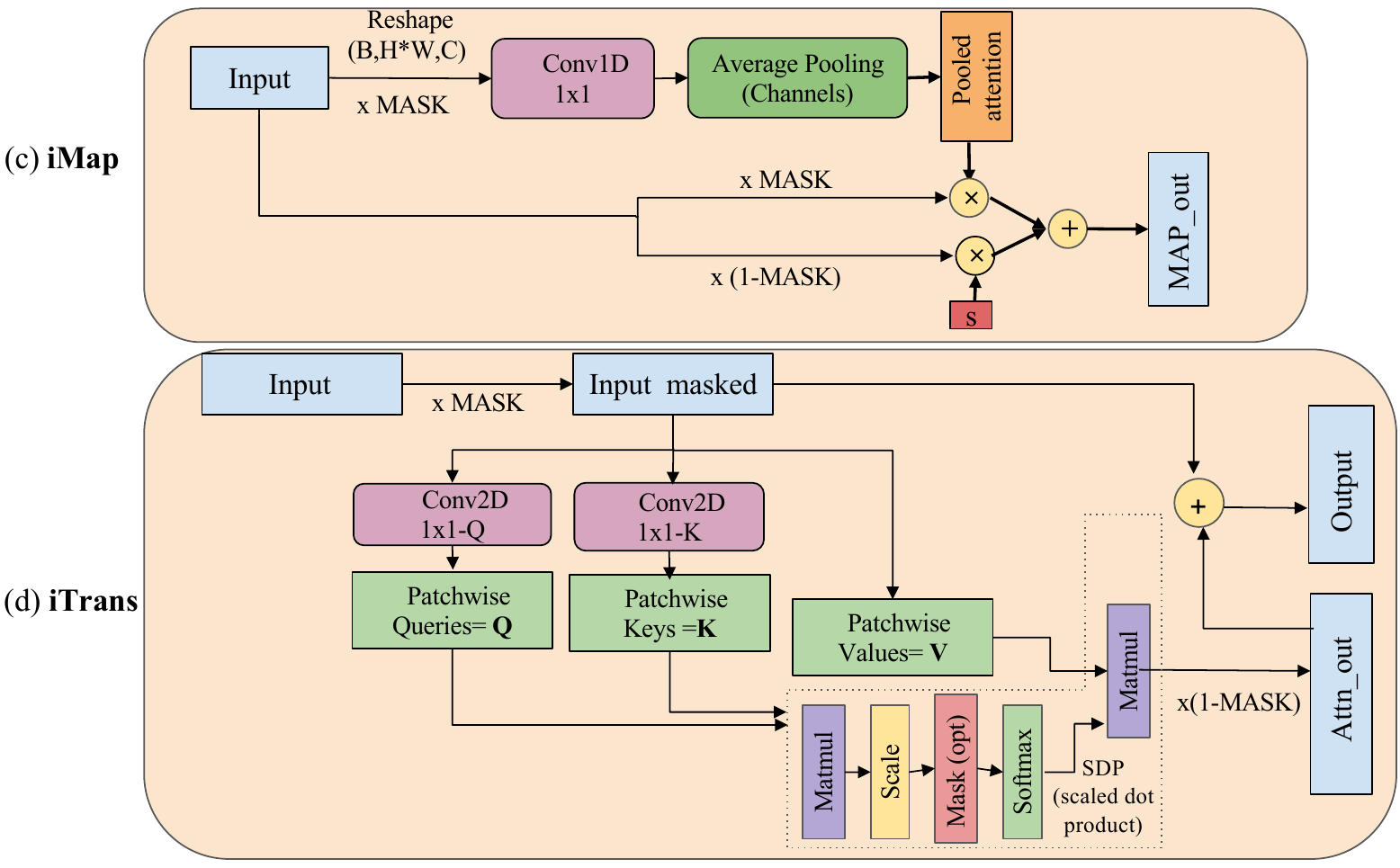}
\end{center}
   \caption{\small (a) Neural architecture of regular flow-based generative models. (b) Neural architecture design of the proposed attention flow model (AttnFlow), which aims at inserting invertible map-based (iMap) attention and transformer-based (iTrans) attention to regular flow-based generative models. $\mathbf{x}, \mathbf{h}_i, \mathbf{z}$ indicates the data, latent variable and intermediate coding respectively. (c) Detailed design of iMap. $B, H, W ,C $ indicate batch size, image hight, width, and channel number respectively. $\times$ MASK represents the masking operation that applies 3D checkboard mask to the input. Conv2D $1 \times 1$ is an invertible 2D convolution. $s$ is a learnable scale. Finally averaged pooled features are fed with learnable parameters into MAP$_\text{out}$ that is sigmoid function. (d) Detailed design of iTrans. MASK indicates the 3D checkboard masking, and Mask(opt) is optional.
 }
\label{fig:attnflow}
\vspace{-0.05cm}
\end{figure*}

The proposed attention flow model (AttnFlow) aims to insert invertible map-based (iMap) and transformer-based (iTrans) attentions to regular flow-based generative models. Fig.(\ref{fig:attnflow}) (a) (b) show the neural architecture design of regular flow-based generative models and the proposed AttnFlow respectively. As shown in Fig.(\ref{fig:attnflow}) (b), the invertible attention modules (i.e., iMap and iTrans) can be stacked on the affine coupling layers. It is also possible to add the attention modules at any other positions, such as before invertible $1 \times 1$ convolution, or actnorm. The detailed architectures of the proposed iMap and iTrans are illustrated in Fig. (\ref{fig:attnflow}) (c) (d) respectively. Their designs follow the conceptual graphs of forward and inverse propagation of the proposed Map-based and Trans-based attention mechanisms that are shown in Fig.(3) of the major paper. In particular,  both of the iMap and iTrans modules apply the 3D checkboard mask in order to make the proposed attention invertible. After the 3D masking,  the iMap attention further applies map-based transformations (i.e., 1D convolution and average pooling) for the first-order attention learning. By comparison, the iTrans attention aims at learning the second-order correlations among the flow feature maps with a scaled dot-product over two different transformations of inputs, which are obtained by two 2D convolutions, respectively. More architecture details of the proposed AttnFlow-iMap and AttnFlow-iTrans are shown in Fig.(\ref{fig:attnflow}) (c) (d). 

% \clearpage

\section{Detailed Experimental Setup}

We used MNIST \cite{lecun1998gradient}, CIFAR10 \cite{krizhevsky2009learning}, CelebA \cite{liu2015deep} and Cityscapes \cite{Cordts2016Cityscapes} datasets to evaluate the proposed AttnFlows in the main paper. MNIST is a dataset of 70,000 small square $28 \times 28$ pixel grayscale images of handwritten single digits between 0 and 9. Following most of the generative modeling works such as \cite{kingma2018glow,mahajan2020normalizing}, we use the whole dataset from the real set to train our AttnFlows and the competing methods. 
CIFAR10 dataset is comprised of 60,000 $32 \times 32$ pixel color images of objects from 10 classes, such as frogs, birds, cats, ships, airplanes, etc. To train the proposed AttnFlows and its competitors, we also utilize the whole dataset for the real data.
We additionally evaluate the proposed cAttnFlows for face super-resolution (8$\times$) using 5000 $160 \times 160$ images from the test split of the CelebA dataset. On CelebA, we use the full train split of CelebA for the training high-resolution image set. Following \cite{lugmayr2020srflow}, we apply a bicubic kernel to down-scale those selected images into $20 \times 20$ low-resolution images. We use 162770 training images following the same setup as \cite{lugmayr2020srflow} for the train-test split. We also use Cityscapes  \cite{Cordts2016Cityscapes} to evaluate the proposed cAttnFlows.
Each instance of this dataset is a $256 \times 256$ picture of a street scene that is segmented into objects of 30 different classes, e.g., road, sky, buildings, cars, and pedestrians. 5000 of these images come with fine per-pixel class annotations of the image, and this is commonly called as segmentation masks. We employ the data splits provided by the original dataset (2975 training and 500 validation images), and train different models to generate street-scene images conditioned on their segmentation masks.

\begin{figure*}[h!]
\centering
\scriptsize
\resizebox{1\linewidth}{!}{%
\begin{tabular}{cc}
% \toprule
\includegraphics[width=0.45\linewidth]{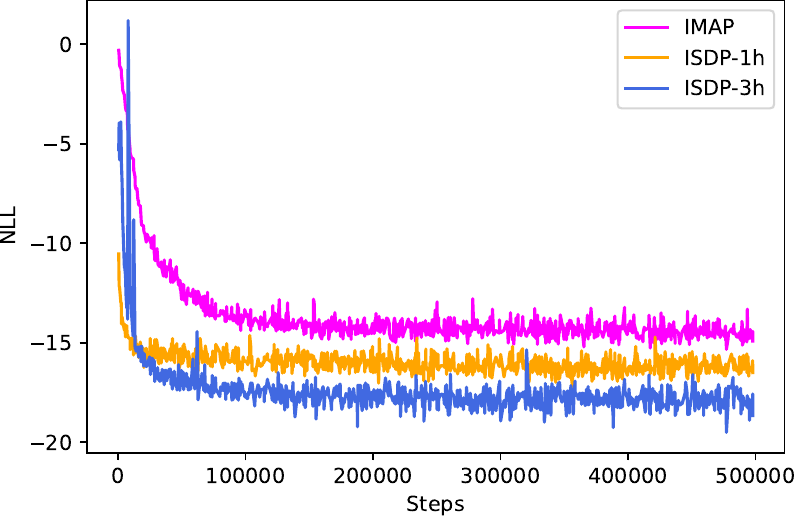} &
\includegraphics[width=0.45\linewidth]{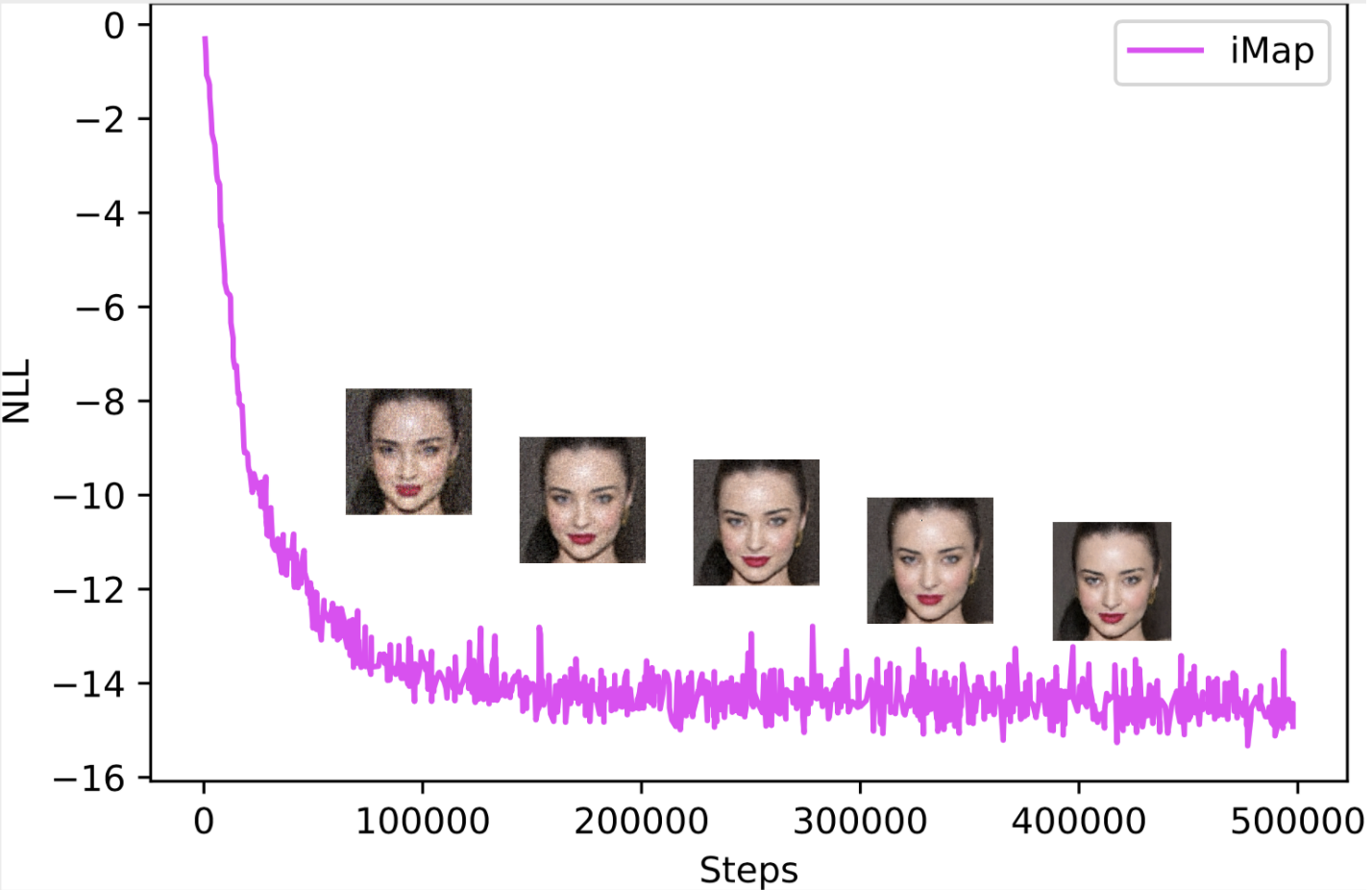}\\

(a) Training Curves  & (b) Generated sample change of AttnFlow-iMap \\

\includegraphics[width=0.45\linewidth]{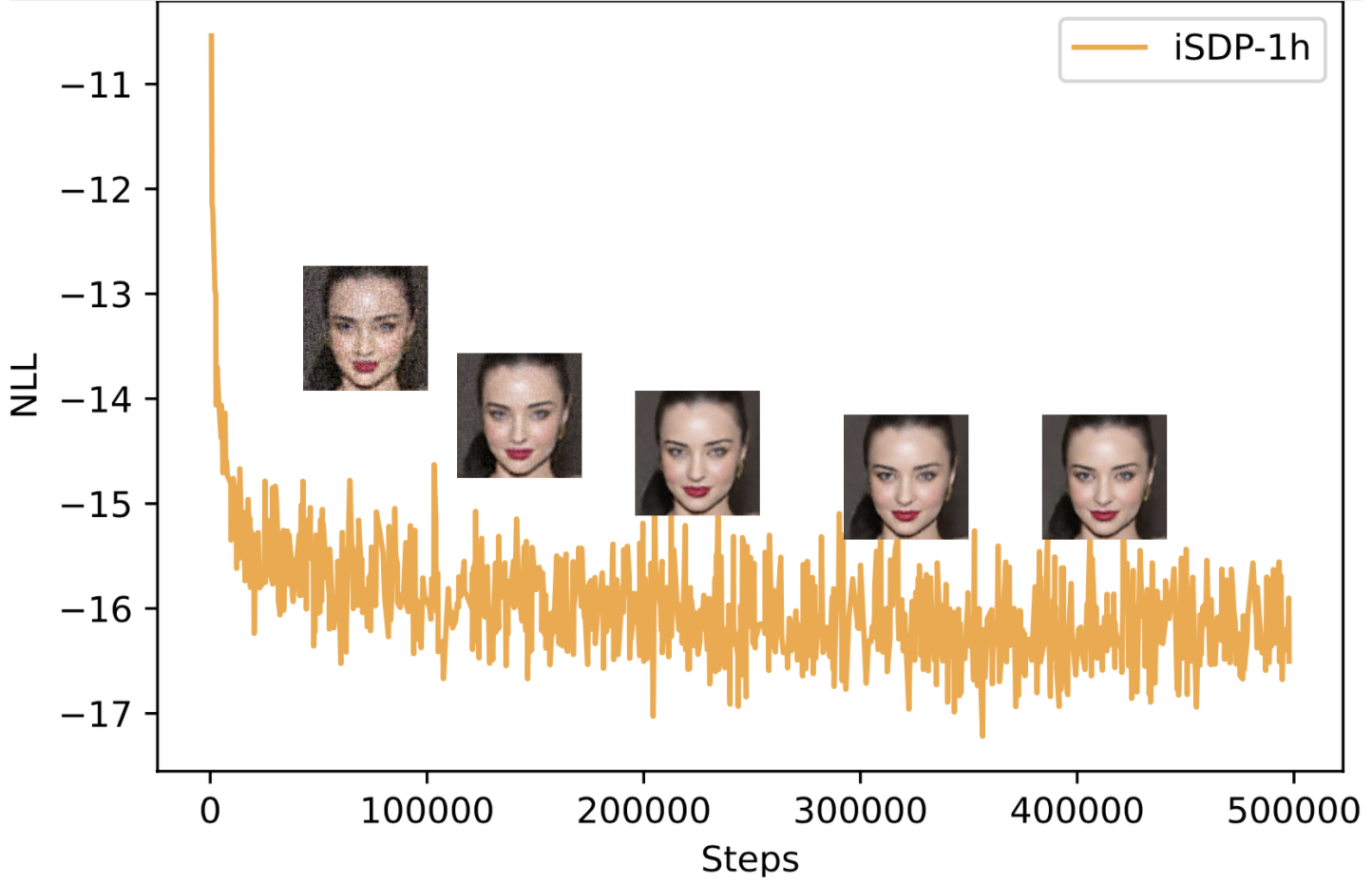} &
\includegraphics[width=0.45\linewidth]{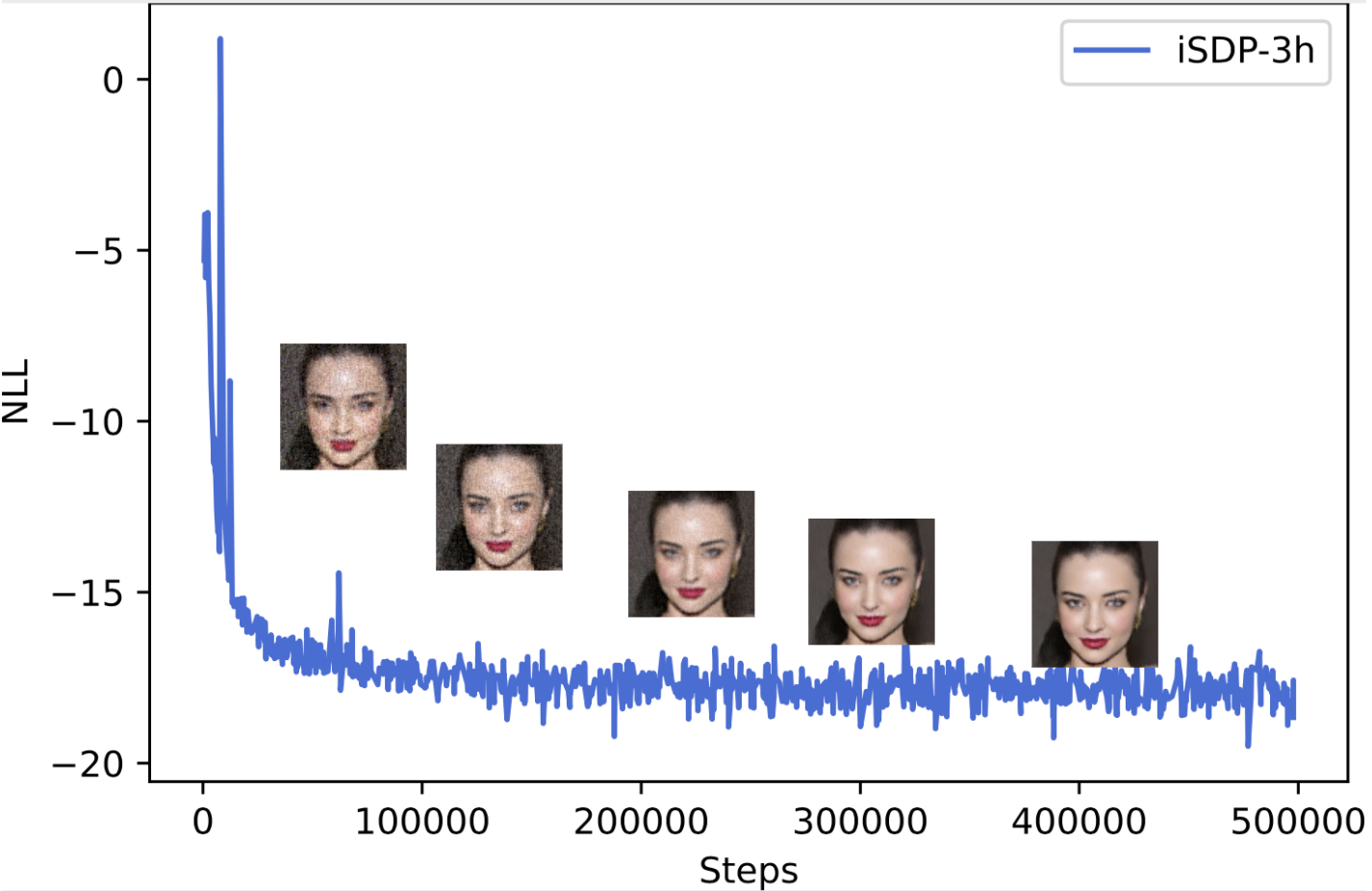}\\

(c) Generated sample change of AttnFlow-iTrans (1 head)  & (d) Generated sample change of AttnFlow-iTrans (3 heads) \\

\end{tabular}
}
\caption{\small (a) Training curves of the proposed AttnFlow-iMap and AttnFlow-iTrans (with 1, 3 head(s)) on CelebA. The x axis corresponds to the training iterations, and the y axis indicates the negative log-likelihood (NLL) values. (b-d) Generated sample change along the training process of AttnFlow-iMap and AttnFlow-iTrans (with 1, 3 head(s)), showing that the generated sample keeps improving the quality until the model converges.}
\label{fig:trainingcurves}
% \vspace{-0.495cm}
\end{figure*}

\begin{figure*}[t]
\centering
\scriptsize
\resizebox{0.8\linewidth}{!}{%
\begin{tabular}{cc}
% \toprule
\includegraphics[width=0.4\linewidth]{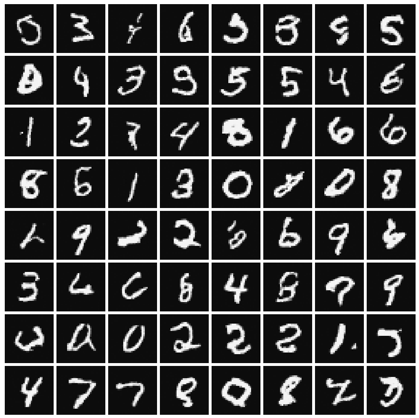} &
\includegraphics[width=0.4\linewidth]{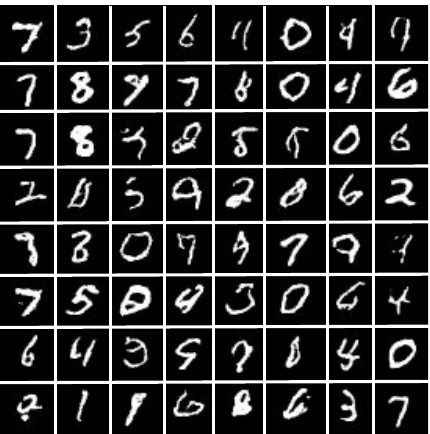}\\

(a) mARFlow \cite{mahajan2020normalizing}  & (b) Proposed \textit{AttnFlow-iMap} \\

\includegraphics[width=0.4\linewidth]{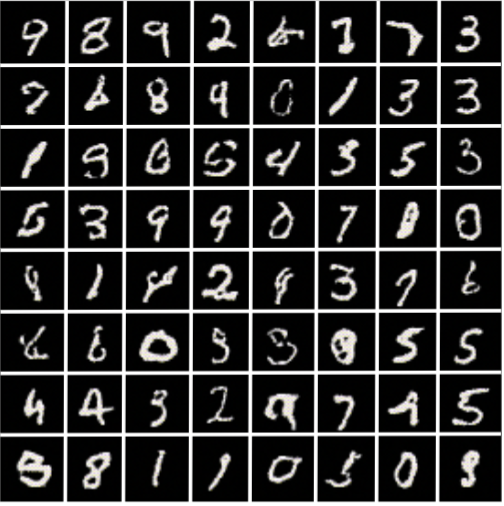} &
\includegraphics[width=0.4\linewidth]{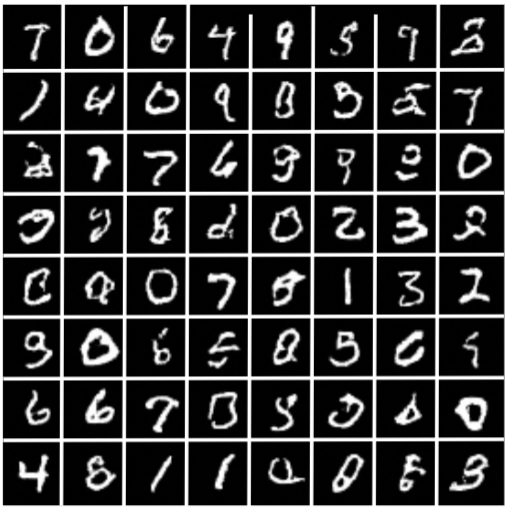}\\

(c) Proposed \textit{AttnFlow-iTrans} (1 head)  & (d) Proposed \textit{AttnFlow-iTrans} (3 heads) \\

\end{tabular}
}
\caption{ \small Random samples generated by the proposed AttnFlows and the state-of-the-art flows on the MNIST dataset.}
\label{fig:mnist_sample}
% \vspace{-0.495cm}
\end{figure*}

\section{Training Details and Curves}

A single TITAN-RTX GPU (24GB) is used to train each of the proposed AttnFlows/cAttnFlows. Specially, the batch size\footnote{Regarding the hyperparmeter setup, we use Adam with a learning rate of $8 \times 10^{-4}$, as done in \cite{mahajan2020normalizing, kingma2018glow}.} is set to 32 for the training on them on both of MNIST and CIFAR10, as done in \cite{mahajan2020normalizing, kingma2018glow}. 
The proposed AttnFlow-iMap and AttnFlow-iTrans models (L=3, K=2) are trained for around 2 days and 3 days on MNIST, and they (L=3,K=4) are trained for 3 days and 5 days on CIFAR10. 
The number of iterations for convergence are 100k iterations and 90k iterations for MNIST and CIFAR10 respectively.
The proposed cAttnFlow-iMap (L=1,K=1) is trained for 1.5 days, and cAttnFlow-iTrans is trained for 2 days on the CelebA datasets.
A batch size is fixed as 16 for all the models on CelebA. The number of steps for convergence are 500k iterations for all the models.
On the Cityscapes dataset, cAttnFlow-iMap (L=2,K=8) is trained for 2 days, and cAttnFlow-iTrans is trained for 3 days. 
A batch size is set to 1 for all the models due to the memory limit. The number of steps for convergence are 200k iterations for all the models on Cityscapes.
Besides, we further compare our AttnFlows-iMap/iTrans against mARFlow \cite{mahajan2020normalizing} and SRFlow \cite{lugmayr2020srflow} (our two main backbones with the same levels and steps as those of ours) in terms of training time, training epochs and test time (per image) in Table \ref{table:comparison}. The results show that the proposed AttnFlows and the competing methods are trained at the same epochs. Their training and inference time are relatively comparable.

\begin{table}[t]
\vspace{-0.2cm}
\begin{center}
\scriptsize
\begin{tabular} {|l|c|c|c|c|}
\hline
Method & Train time & Train epoch & Test time\\
\hline
mARFlow (CIFAR) & 2.5 GPU days
  & ~100 & 0.48s\\
AttnFlow-iMap (CIFAR) & 3 GPU days & ~100 & 0.50s \\
AttnFlow-iTrans (CIFAR) & 5 GPU days & ~100 & 0.61s \\
\hline
SRFlow (CelebA) & 1 GPU day & ~20 & 0.077s \\
AttnFlow-iMap (CelebA) & 1.5 GPU days & ~20 & 0.104s \\
AttnFlow-iTrans (CelebA) & 2 GPU days & ~20 & 0.317s \\
\hline
\end{tabular}
\end{center}
\vspace{-0.2cm}
\caption{\footnotesize Comparison of the proposed AttnFlow-iMap and AttnFlow-iTrans against their corresponding backbones mARFlow/SRFlow.}
\label{table:comparison}
\vspace{-0.6cm}
\end{table}

Fig.(\ref{fig:trainingcurves}) (a) shows the training curves of the proposed AttnFlow-iMap and AttnFlow-iTrans on CelebA. It reflects that the proposed AttnFlows can be trained smoothly for a good convergence in terms of the negative log-likelihood (NLL) loss. In addition, Fig.(\ref{fig:trainingcurves}) (b)(c)(d) demonstrate the sample change along the training processes of the proposed AttnFlows. They show that the generated samples can keep improving the quality until the convergence.

\section{More Results for MNIST, CIFAR10, CelebA and Cityscapes}

Fig.(\ref{fig:mnist_sample}), Fig.(\ref{fig:cifar10_samp}), Fig.(\ref{fig:celeba_sample}) and Fig.(\ref{fig:cityscapes}) show more visual results of the proposed AttnFlows and the competing methods on MNIST, CIFAR10, CelebA and Cityscapes respectively. From the results, we can observe that the generated samples of our proposed AttnFlows are highly competitive, and some are more visually pleasing compared to the competing methods.

\begin{figure*}[h!]
\centering
\scriptsize
\resizebox{1\linewidth}{!}{%
\begin{tabular}{cc}
% \toprule
\includegraphics[width=0.4\linewidth]{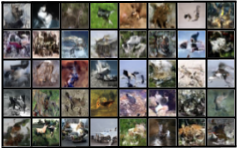} &  \includegraphics[width=0.4\linewidth]{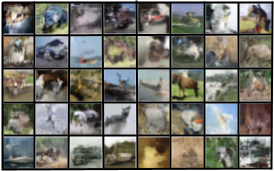}\\

(a) Flow++ \cite{ho2019flow++} (3.29 bits/dim) & (b) mARFlow \cite{mahajan2020normalizing} (3.24 bits/dim, 41.9 FID)

\\

 \includegraphics[width=0.4\linewidth]{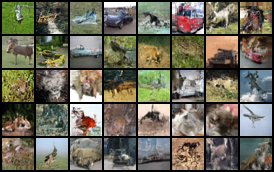} &
\includegraphics[width=0.4\linewidth]{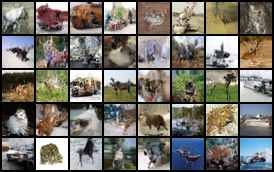}\\
(c) \textit{AttnFlow-iMap} (3.216 bits/dim, 33.6 FID) & (d) \textit{AttnFlow-iTrans} (3.217 bits/dim, 33.8 FID) \\

\end{tabular}
}
\caption{\small Random samples generated by the proposed AttnFlows and the state-of-the-art flows on the CIFAR10 dataset. }
\label{fig:cifar10_samp}
% \vspace{-0.3cm}
\end{figure*}

\begin{figure*}[h!]
    \centering%
    \newcommand{\size}{0.15}%
    \newcommand{\img}[1]{%
    \includegraphics[width=\size\linewidth]{figures/sotaCMbic8/#1/lq}~~~~%
    \includegraphics[width=\size\linewidth]{figures/sotaCMbic8/#1/ESRGANn23px2e-1.jpg}~~%
    \includegraphics[width=\size\linewidth]{figures/sotaCMbic8/#1/srflow.png}~~%
    \includegraphics[width=\size\linewidth]{figures/sotaCMbic8/#1/imap.png}%
    \includegraphics[width=\size\linewidth]{figures/sotaCMbic8/#1/isdp.png}%

    }%
    \img{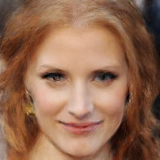}
    \img{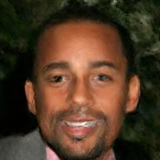}
    \img{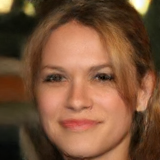}
    \img{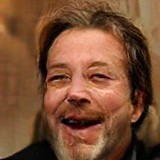}
    \resizebox{0.8\linewidth}{!}{%
    \begin{tabular}{ C{3.05cm} C{2.5cm} C{2.5cm} C{2.5cm} C{2.5cm} }
        Low-resolution input  &
        ESRGAN~\cite{wang2018esrgan}  & SRFlow~\cite{lugmayr2020srflow} & \textit{cAttnFlow-iMap} &  \textit{cAttnFlow-iTrans}
    \end{tabular}}%
    \caption{\small Super-resolved samples of the proposed cAttnFlows and the state-of-the-art models for $8\times$ face SR on the CelebA dataset.}
    \label{fig:celeba_sample}
    \vspace{-0.3cm}
\end{figure*}

\begin{figure*}[h!]
\centering
\scriptsize
\resizebox{0.8\linewidth}{!}{%
\begin{tabular}{cc}
% \toprule
\includegraphics[width=0.32\linewidth]{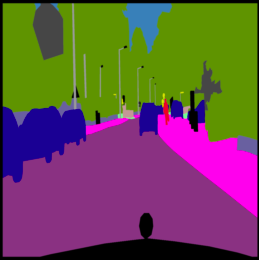} &
\includegraphics[width=0.32\linewidth]{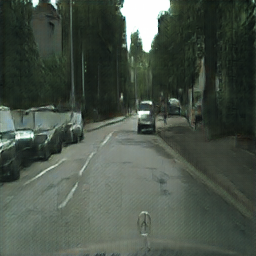}\\

(a) Input  & (b) Pix2PixGAN \cite{isola2017image}\\

\includegraphics[width=0.32\linewidth]{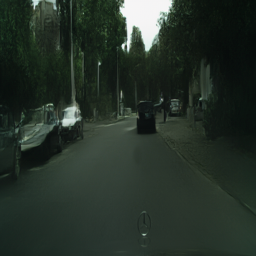} &
\includegraphics[width=0.32\linewidth]{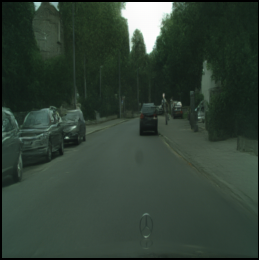}\\

(c) Dual-Glow \cite{sun2019dual}  & (d) Full-Glow \cite{sorkhei2020full} \\

\includegraphics[width=0.32\linewidth]{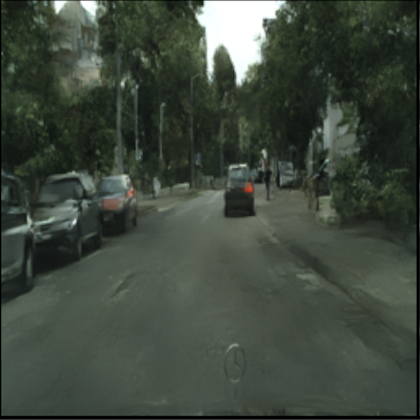} &
\includegraphics[width=0.32\linewidth]{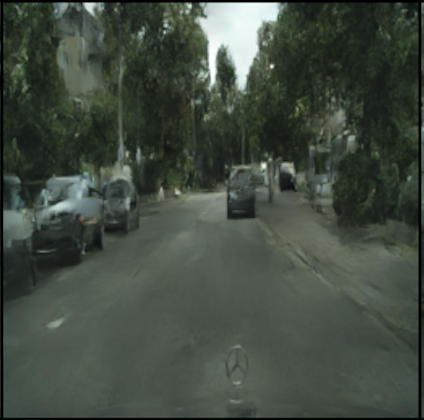}\\
(e) Proposed \textit{cAttnFlow-iMap} & (f) Proposed \textit{cAttnFlow-iTrans} \\

\end{tabular}
}
\caption{\small 
Generated samples of the proposed cAttnFlows and the state-of-the-art models for image translation on the Cityscapes dataset. The competing methods and ours are conditioned on the semantic segmentation labels (a) to synthesize the RGB images with the resolution being of $256 \times 256$.
}
\label{fig:cityscapes}
\end{figure*}

\section{More Ablation Study}

\begin{figure*}[h!]
\begin{center}
\vspace{-0.3cm}
        \includegraphics[width=0.51\linewidth]{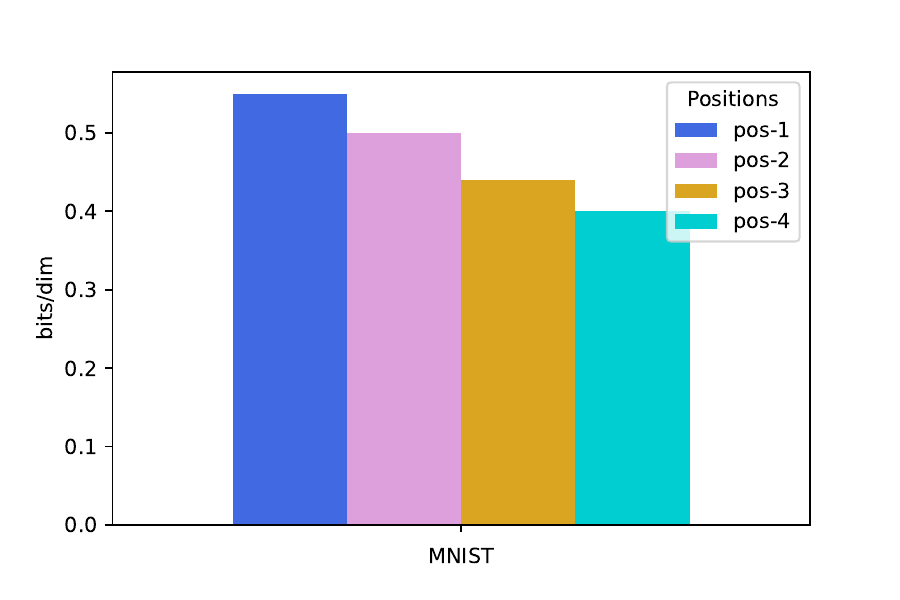} 
      \includegraphics[width=0.48\linewidth]{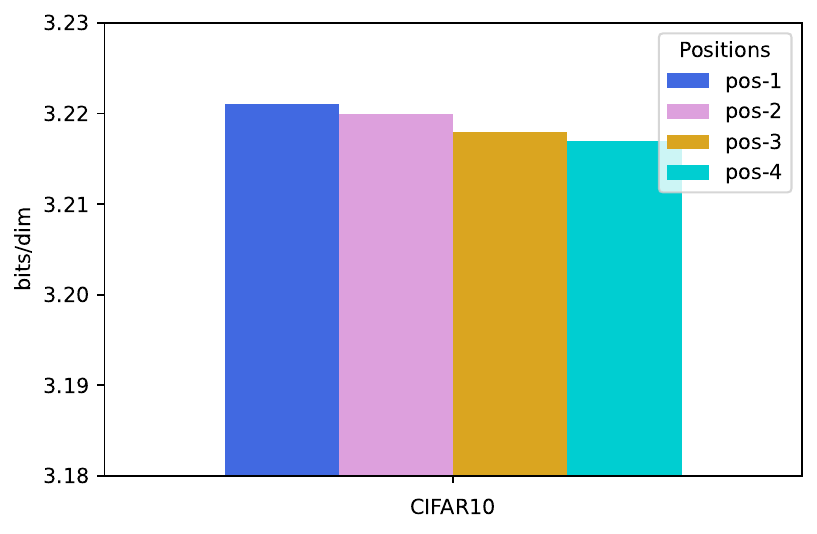}  \\
            
        \includegraphics[width=0.34\linewidth]{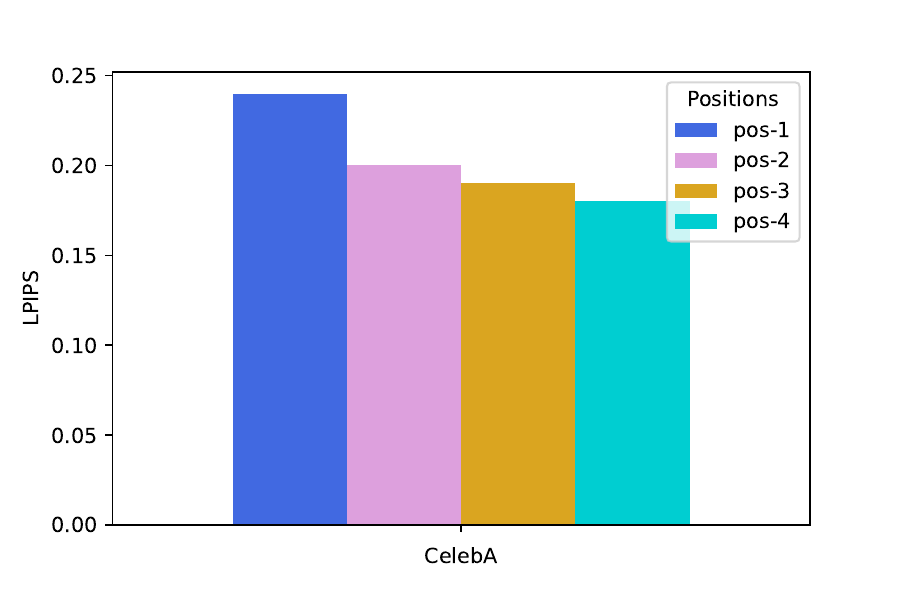}
      \includegraphics[width=0.32\linewidth]{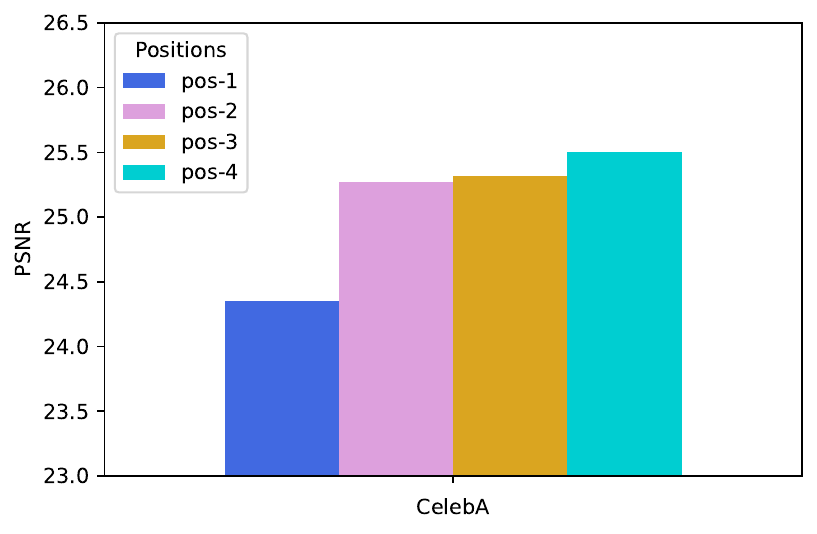}
         \includegraphics[width=0.32\linewidth]{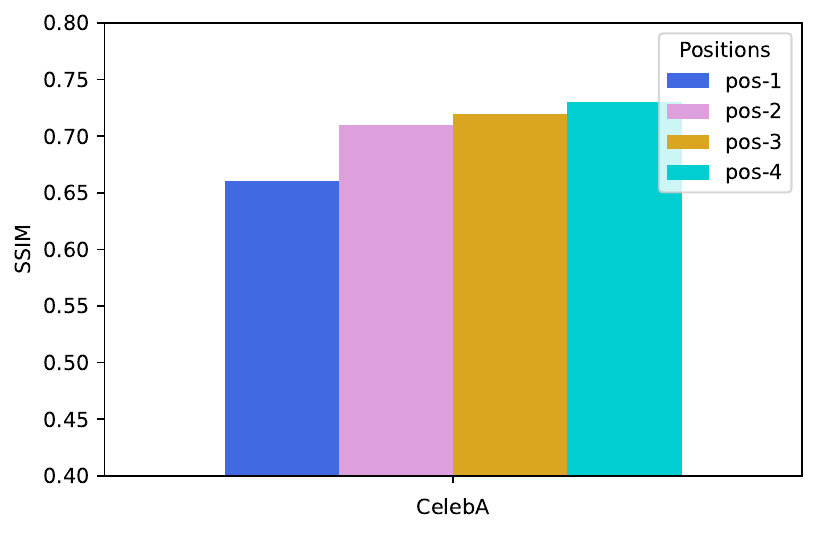} \\
         
\end{center}
% \vspace{-0.2cm}
   \caption{\small (Better/more visualisation for Fig.(5) in the major paper) 
   Ablation studies of the proposed attentions on different positions in the flow layers (pos-1: before actnorm, pos-2: after actnorm, pos-3: after permutation, pos-4: after coupling layer) on the MNIST, CIFAR10 and CelebA datasets. The Bits/dims metric is employed for MNIST and CIFAR10 (top), and LPIPS, PSNR and SSIM scores are reported on CelebA (bottom).
   }
\label{fig:abation_position}
% \vspace{-0.5cm}
\end{figure*}

\begin{figure*}[h!]
\begin{center}
\vspace{-0.3cm}

        \includegraphics[width=0.49\linewidth]{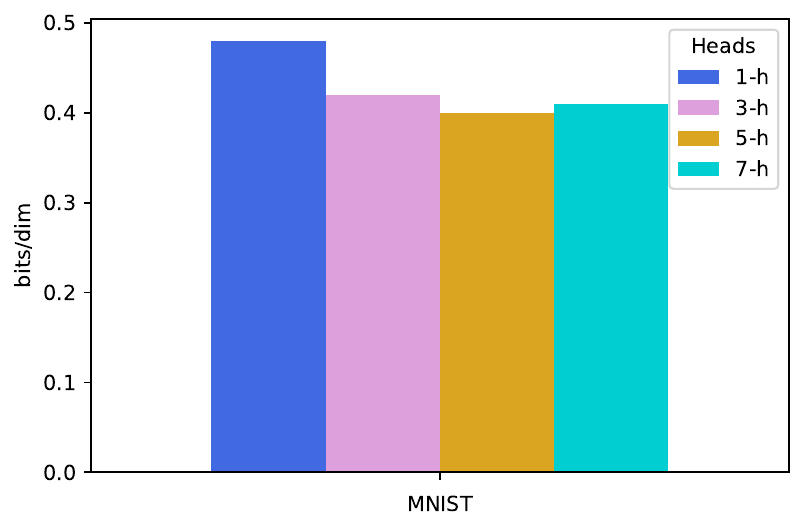}    
         \includegraphics[width=0.49\linewidth]{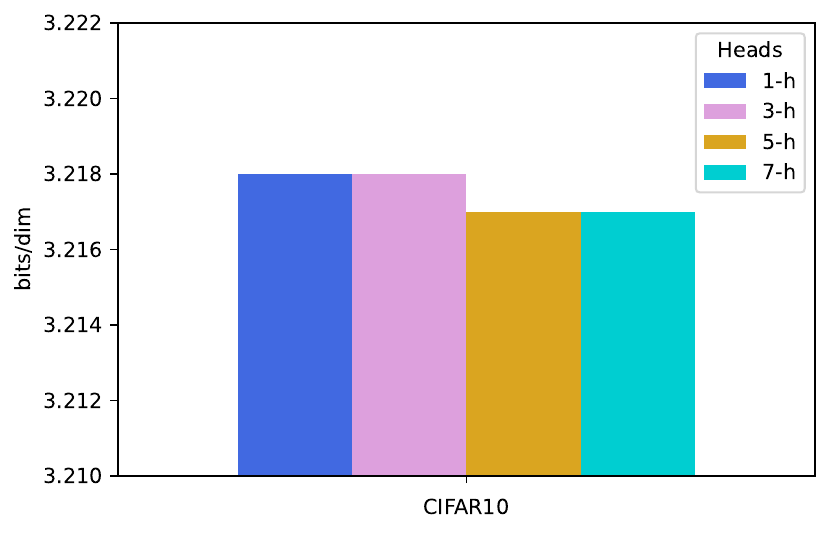}\\
            \includegraphics[width=0.32\linewidth]{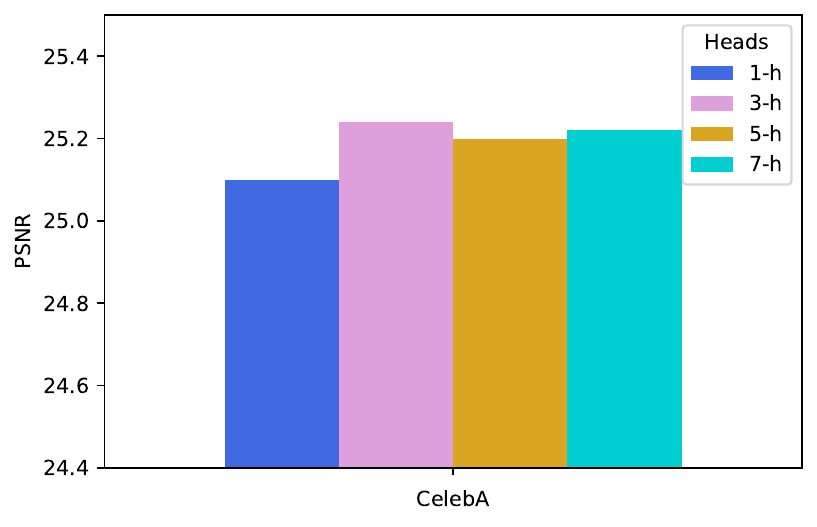}
  \includegraphics[width=0.34\linewidth]{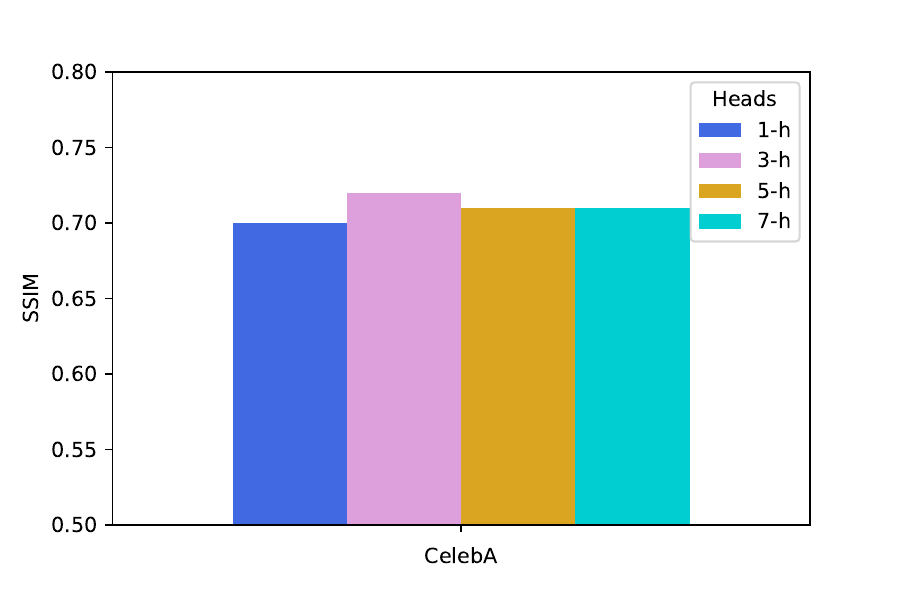}
   \includegraphics[width=0.32\linewidth]{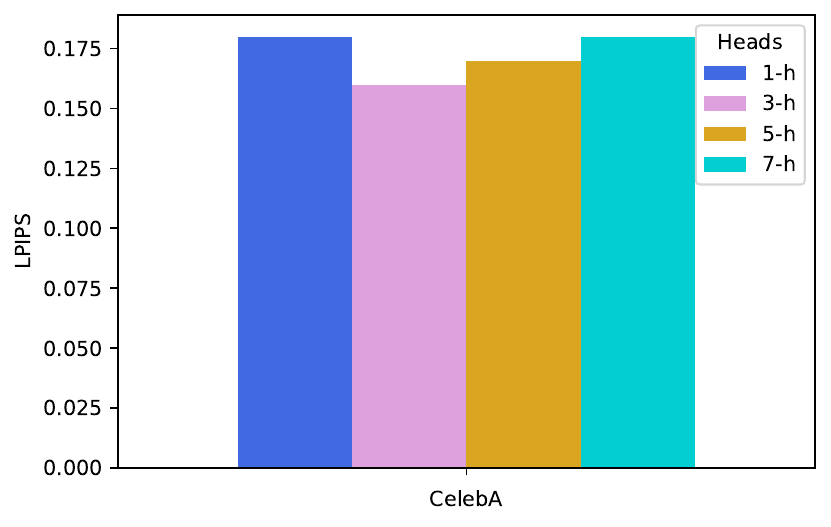}
\end{center}
% \vspace{-0.2cm}
   \caption{\small (Better/more visualisation for Fig.(5) in the major paper) 
   Ablation studies of the proposed attentions on different number of attention heads for the proposed iTrans attention (1h: 1head, 3h: 3 heads, 5h: 5 heads, 7h: 7 heads) on the MNIST, CIFAR10 and CelebA datasets. The Bits/dims metric is employed for MNIST and CIFAR10 (top), and LPIPS, PSNR and SSIM scores are reported on CelebA (bottom).  
   }

\label{fig:abation_head}
% \vspace{-0.5cm}
\end{figure*}

For the ablation study on the proposed AttnFlows, better/more visualizations of the major paper's Fig.(5) are shown in Fig.(\ref{fig:abation_position}) and Fig.(\ref{fig:abation_head}). Additionally, Fig.(\ref{fig:abation_head}) includes the ablation study on different head numbers of the proposed AttnFlow-iTrans for CelebA. As shown in Fig.(\ref{fig:abation_position}), inserting the attention modules after each coupling layer generally works the best in the most of cases. Besides, Fig.(\ref{fig:abation_head}) shows that using 5 heads performs the best on the MNIST and CIFAR10 datasets, while employing 3 heads works the best on CelebA. This implies that using 3 or 5 heads is sufficient for the proposed AttnFlow-iTrans on the three employed datasets.

\section{Pseudo Code of Proposed AttnFlows}

\begin{figure*}[h!]
\begin{center}
\vspace{-0.1cm}
 \includegraphics[width=0.7\linewidth,height=10.5cm]{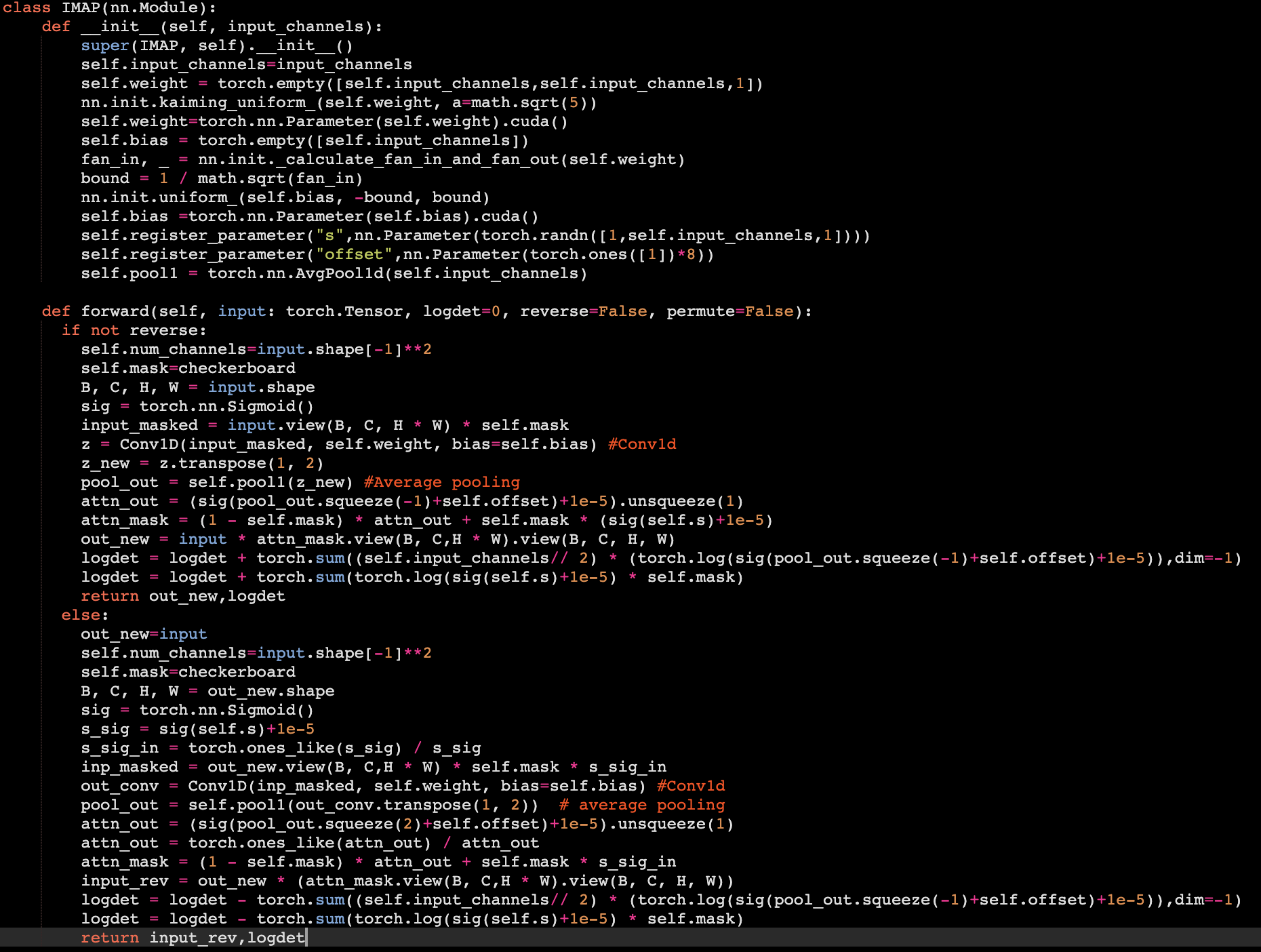}
  \includegraphics[width=0.7\linewidth,height=10.5cm]{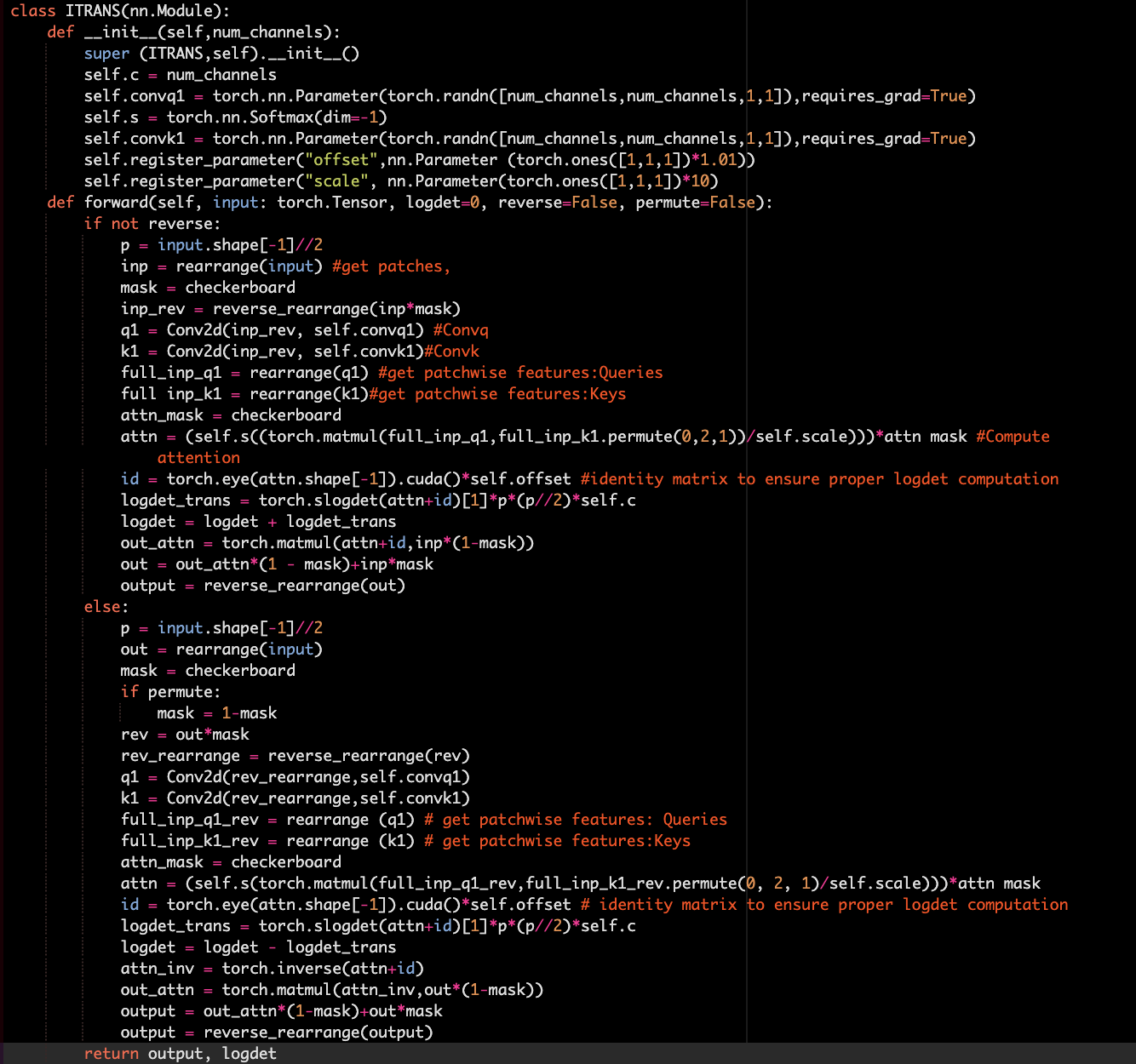}

\end{center}
% \vspace{-0.2cm}
   \caption{\small Pseudo code of the proposed AttnFlow-iMap and AttnFlow-iTrans.}
\label{fig:pseudocode}
\vspace{-0.5cm}
\end{figure*}

The proposed AttnFlow-iMap and AttnFlow-iTrans are built upon the off-the-shelf flow models. Therefore, the major new implementation is on the proposed iMap and iTrans modules. The pseudo codes for their PyTorch implementation are illustrated in Fig.(\ref{fig:pseudocode}).

\section{Further Remarks for the Future Work}
For the proposed AttnFlow-iTrans, we introduce a masked version of scaled dot-product, where the introduced masking can serve as a transformation (i.e., binary pattern generation). To improve the generalization capability, we could further apply a $1 \times 1$ 2D convolution to the value $\mathbf{V}$, as done on the query $\mathbf{Q}$ and the key $\mathbf{K}$. Also, we exploit the multi-head attention mechanism for the AttnFlow-iTrans. To ease the computation on Jacobian determinant of iTrans, we choose to perform a summation instead of the commonly-used concatenation in conventional transformers over the resulting attended features from multi-heads. Following this work, we will be making more comprehensive study on the full exploitation of the multi-head attention scheme.
Besides, as discussed in the main paper, it is non-trivial to apply the proposed AttnFlows to deeper flows, such as the full SRFlow model that contains more flow levels. We study that it is mainly because the proposed attentions' inverse and Jacobian determinant computations are often numerically unstable when meeting deeper flows. This is roughly matched with the discovery in \cite{dong2021attention}, which finds pure attentions typically lose rank doubly exponentially with the network depth. Inspired by \cite{dong2021attention}, a natural solution is to apply residual learning that is capable of addressing the deep attention problem. In addition, the comprehensive evaluations over the four used datasets show that the proposed AttnFlow-iMap sometimes outperforms AttnFlow-iTrans, while the former also performs worse in some cases. Hence, it is valuable to optimize the aggregation of the two complementary types of attention (i.e., first- and second-order attentions) for real-world scenarios. To this end, one of the most promising directions is to exploit neural architecture search algorithms over them.

\end{document}